\pdfoutput=1

\documentclass[11pt]{article}

\usepackage[]{paper}
\usepackage{times}
\usepackage{latexsym}
\usepackage{booktabs}
\usepackage{amsmath}
\usepackage{amssymb}
\usepackage{caption}
\usepackage{subcaption}
\usepackage{siunitx}
\usepackage{nameref}

\usepackage[T1]{fontenc}

\usepackage[utf8]{inputenc}

\usepackage{microtype}
\usepackage{todonotes}

\title{How Far Can It Go? On Intrinsic Gender Bias Mitigation for Text Classification}

\author{Ewoenam Kwaku Tokpo$^1$, Pieter Delobelle$^{2,3,5}$, Bettina Berendt$^{2,3,4,5}$ \and Toon Calders$^1$ \\
$^1$ {Department of Computer Science, University of Antwerp, Belgium}\\
$^2$ Department of Computer Science, KU Leuven, Belgium \\
$^3$ Leuven.AI Institute for Artificial Intelligence, Belgium\\
$^4$ Faculty of Electrical Engineering and Computer Science, TU Berlin, Germany \\
$^5$ Weizenbaum Institute, Germany\\
}

\begin{document}
\maketitle
\begin{abstract}
To mitigate gender bias in contextualized language models, different intrinsic mitigation strategies have been proposed, alongside many bias metrics.
Considering that the end use of these language models is for downstream tasks like text classification, it is important to understand how these intrinsic bias mitigation strategies actually translate to fairness in downstream tasks and the extent of this.
In this work, we design a probe to investigate the effects that some of the major intrinsic gender bias mitigation strategies have on downstream text classification tasks. 
We discover that instead of resolving gender bias, intrinsic mitigation techniques and metrics are able to hide it in such a way that significant gender information is retained in the embeddings.
Furthermore, we show that each mitigation technique is able to hide the bias from some of the intrinsic bias measures but not all, and each intrinsic bias measure can be fooled by some mitigation techniques, but not all.
We confirm experimentally, that none of the intrinsic mitigation techniques used without any other fairness intervention is able to consistently impact extrinsic bias.
We recommend that intrinsic bias mitigation techniques should be combined with other fairness interventions for downstream tasks.
\end{abstract}

\section{Introduction}

The use of pretrained language models has seen a surge in popularity as a result of state-of-the-art performances that have been achieved with these models on various tasks. Consequently, there has been a growing interest in exploring how gender bias pertains in these models \citep{garrido2021survey}.
Pretrained language models are used in two distinct phases: the \emph{pretraining phase} and the \emph{finetuning phase}. The pretraining phase typically involves training a model on a generic task such as masked language modeling on a diverse set of text corpora. In the finetuning phase, the pretrained model can be adapted for a specific task, such as sentiment analysis, by training on a domain-specific corpus.
Owing to the unique way of using pretrained models, bias generally manifests in two forms: \emph{intrinsic bias} and \emph{extrinsic bias}. Intrinsic bias refers to bias that inherently exists in pretrained language models whereas extrinsic bias is used to refer to bias that exists in downstream models that are based on the pretrained model.
Since the success of downstream NLP tasks has mostly been dependent on pretrained models, it is intuitive to assume that bias in intrinsic models will 
compromise fairness in downstream tasks.
Only recently have more in-depth examinations been done to investigate this assumption \citep{steed-etal-2022-upstream,orgad-etal-2022-gender, kurita-etal-2019-measuring}.
However, conclusions have been inconsistent and the confounding effects of bias mitigation techniques remain unknown.

The main focus of this work is to investigate the impact of intrinsic bias on extrinsic fairness 
and if techniques to mitigate intrinsic bias actually resolve bias or only mask it.
We develop a probe to uncover intrinsic bias by determining the amount of gender information in a word embedding using a classifier.
Bearing in mind how abstractly and improperly intrinsic bias has been defined \citep{Blodgett2020Language}, coupled with the discovery that results from different metrics for intrinsic bias in many cases do not correlate \citep{delobelle-etal-2022-measuring}, we find this probe effective as an extra step in evaluating the efficacy of these mitigation strategies.
 We realize from this study that how intrinsic bias has been measured and the choice of bias mitigation strategies explored by some existing works have not been ideal.
We further use this probe to investigate if some proposed mitigation strategies superficially conceal bias.

In this work, when we refer to bias in a language model, we mean stereotyping bias as defined by \citet{garrido2021survey} as \emph{``the undesired variation of the [association] of certain words in that language model according to certain prior words in a given domain''}.
We focus our experiments on gender bias for two primary reasons: its intuitive nature making it easy to analyze and discuss, and the accessibility of resources and datasets regarding gender. For the same reason, we narrow our definition of bias in our experiments and analysis to binary gender bias\footnote{See \nameref{sec:ethical considerations}.}. 
This paper considers the primary goal of mitigating intrinsic bias to ensure fairness in downstream tasks.

We consider one case of binary classification and one multiclass classification case, all on English language corpora using the BERT-large \citep{devlin-etal-2019-bert} and the ALBERT-large \citep{lan2019albert} pretrained models for each task.

In summary, we develop an extensive probe \footnote{We make our code available at
\url{https://github.com/EwoeT/intrinsic-gender-probe}} to uncover intrinsic bias in pertained contextualized language models, and seek to answer three key research questions:
\textbf{RQ1:} Do different intrinsic bias metrics respond differently to different bias mitigation techniques? (\autoref{ss:rq1}).
\textbf{RQ2:} Can intrinsic bias mitigation techniques 
    hide bias instead of resolving it? (\autoref{ss:rq2}).
 \textbf{RQ3:} Do intrinsic bias mitigation techniques in language models improve fairness in downstream tasks? (\autoref{ss:rq3})

\section{Measuring and mitigating bias} 

\begin{figure*}[t]
    \centering
    \includegraphics[width=0.7\linewidth]{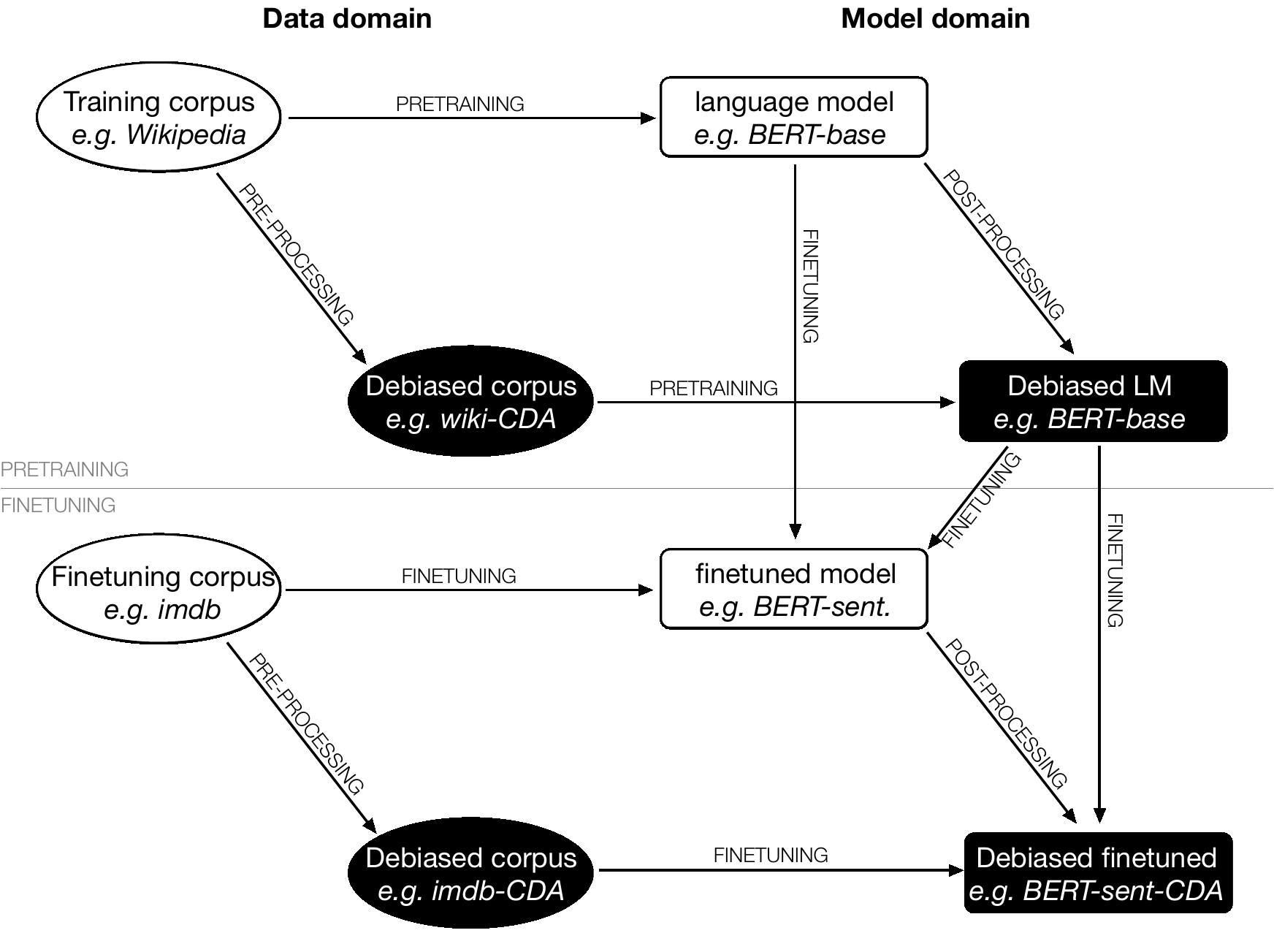}
    \caption{Illustration of where bias mitigation techniques alter a model or training data for both intrinsic (pretrained) and extrinsic (finetuned) biases in models.}
    \label{fig:bias-mitigation}
\end{figure*}

Since many techniques for measuring and mitigating gender bias have been introduced for both intrinsic and extrinsic bias, we only discuss techniques we use in experiments in this paper.

Bias mitigation techniques can be applied to pretrained or finetuned language models, or both. 
\autoref{fig:bias-mitigation} illustrates these interactions with both training settings and this section will discuss both intrinsic (\autoref{ss:mitigating-intrinsic-bias}) and extrinsic (\autoref{ss:mitigating-extrinsic-bias}) mitigation techniques.
Additionally, this section will provide a brief overview of bias measures \footnote{For a more in-depth overview of mainly intrinsic measures, we refer to \citet{delobelle-etal-2022-measuring}.}.

\subsection{Intrinsic bias mitigation techniques}
\label{ss:mitigating-intrinsic-bias}
Intrinsic gender bias mitigation methods target either the pretraining data, 
the pretraining procedure, 
or the pretrained model's output, 
which we refer to as pre-processing, in-processing, and post-processing respectively~\citep{friedler2019comparative}.
We select three popular mitigation methods to represent all three types, namely \emph{Counterfactual Data Augmentation} (CDA), \emph{Context-debias}, and \emph{Sent-debias}.
Notice that these methods create debiased pretrained language models, as is illustrated in \autoref{fig:bias-mitigation}.
These models still need to be finetuned on a downstream task.

\paragraph{CDA pretraining.}\label{para:pretrained_CDA}
The idea behind counterfactual data augmentation \citep{zmigrod-etal-2019-counterfactual, lu2020gender} is to generate a counterfactual for each example in the training corpus by replacing attribute terms with their complimentary equivalent from the other demographic classes. For example, \emph{she} will map to \emph{he} in the case of binary gender.
To mitigate intrinsic bias, this counterfactual augmentation has to be done as a pretraining step.
Since CDA involves retraining the model, it is more resource-intensive compared to Sent-debias and Context-debias.
We use the pretrained CDA models based on BERT and ALBERT from \citet{zari} for our implementation.


\paragraph{Context-debias.} 
\citet{kaneko-bollegala-2021-context} introduce a debiasing method that involves retraining the language model with a constraint to make the embeddings of \emph{stereotype terms} \footnote{We also refer to them as \emph{target words}.} (such as \emph{doctor, nurse}) orthogonal to embeddings of \emph{attribute terms} \footnote{We also refer to them as \emph{identity terms} or \emph{protected words}.} (such as gender pronouns like \emph{she, he} and gender nouns like \emph{woman, man}).
Given the dynamic nature of contextualized word embeddings which causes a word to have different embeddings, they define fixed word embeddings for each attribute token by averaging the contextual embeddings of a word in all sentences it appears in. 
Training is done so that the embeddings of all stereotype terms are made orthogonal to all fixed attribute embeddings. 
They add a regularizer that constrains the debiased embedding to retain as much information by ensuring that they are as similar to the original embeddings as possible despite the orthogonality (\autoref{ss:Context-debias formula}). Context-debias, as well as other in-processing techniques, require a predefined set of attribute and target terms before training which may not be effective for words outside these sets.


\paragraph{Sent-debias.}
\citet{liang-etal-2020-towards} propose a post-processing debiasing method akin to \emph{word embedding debiasing} \citep{bolukbasi2016man} but for contextualized embeddings.
They achieve this by first identifying the bias subspace. 
They extract naturally-occurring sentences from corpora that contain certain attribute terms, generate counterfactuals for each one, and compute the gender subspace based on these sentence templates.
The rest of their algorithm follows the method of hard debiasing (neutralization and equalization) by \citet{bolukbasi2016man} (\autoref{ss:Sent-debias formula}). 
Essentially, this post-hoc approach transforms the embeddings by removing their projections onto the gender subspace.

\subsection{Measuring intrinsic bias} \label{ss:measuring-intrinsic-bias}
To measure intrinsic bias in the pretrained models, we use \emph{SEAT} and \emph{LPBS}. 
We examine these two metrics because of their wide use in related literature.


\paragraph{Sentence Embedding Association Test (SEAT).}
\citet{may-etal-2019-measuring} developed SEAT to quantify intrinsic bias in contextualized language models based on WEAT \citep{Caliskan} which was originally designed for non-contextualized embeddings. 
Although the authors express concern over the efficacy of this metric, it has been widely adopted in various works to quantify bias in contextualized word embeddings. 
We adopt the implementation by \citet{tan2019assessing} that uses word-level contextualized embeddings of attribute and stereotype terms instead of sentence-level embeddings. \citet{delobelle-etal-2022-measuring} show that using word-level embeddings produces more consistent results and is more robust against the effects of template choices.
SEAT is defined as: $
s(X_f, X_m,A, B) = \sum_{x_f \in X_f} s(x_f,A, B) -\sum_{x_m \in X_m} s(x_m,A, B)  
$;
$A$ and $B$ are the respective sets of female and male attribute (identity) terms, whereas $X_f$ and $X_m$ are the sets of female and male stereotypes respectively,
The similarity measure $s(x,A, B)$ gives the association between a stereotype word's embedding and the word embeddings of attributes:
$
s(x,A, B) =  {\frac{1}{|A|}\sum_{a\in A}}\cos\left(x,a\right) -  {\frac{1}{|B|} \sum_{b\in B} \cos \left(x,b\right)} 
$

\paragraph{Log Probability Bias Score (LPBS).}
This template-based approach proposed by \citet{kurita-etal-2019-measuring} quantifies bias using templates containing target words: such as ``\emph{\_ is a doctor.}''. They compute the difference in probability scores that a language model uses to predict words from two respective groups (eg. \emph{female:she, male:he}) to fill in the blank. To account for the effect of prior probabilities of gender words that may skew results, they normalize the results by dividing each prediction by the prior probability of the attribute term. We compute LPBS as:
$
    LPBS = \sum_{x\in X}\sum_{i} |ls(a_i,x) - ls(b_i,x)|
$; 
where  $a_i$ and $b_i$ are equivalent forms of a female and male attribute term respectively, 
$X = X_f \cup X_m$ is the set of all stereotype words,
$
    ls(w,x) = \log (\frac{P(w|x)}{P(w)})$ is the association between an attribute term $w \in A \cup B$ and a target term $x \in X$,
$P(w)$ is the prior probability of $w$, and $P(w|x)$ is the probability of predicting $w$ as the blank term given the presence of $x$ in its context.

\subsection{Extrinsic bias mitigation techniques}\label{ss:mitigating-extrinsic-bias}
We look at two downstream tasks, which are both text classification (Bias-in-Bios \cite{De-Arteaga-et-al-2019-bias} and Jigsaw\footnote{https://www.kaggle.com/c/jigsaw-unintended-bias-in-toxicity-classification/data?select=train.csv})\footnote{See \autoref{sec:Datasets} for dataset description}.
To understand the relationship between intrinsic and extrinsic bias, in addition to these original datasets, we include two pre-processed versions of the datasets using \emph{Attribute scrubbing} and \emph{Attribute swapping}. 
We use these settings to analyze how intrinsic bias is propagated to downstream tasks. 

\paragraph{Attribute scrubbing.}\label{para:Attribute scrubbing}
Attribute features, which are identity terms in an NLP task, are completely removed from text instances in the training data \citep{prost-etal-2019-debiasing, de2019bias}. 
This approach may seem intuitive, particularly to mitigate explicit bias, but we suggest that deleting tokens in NLP tasks can be tricky since it could change the syntactic and grammatical structure of the text, leading to out-of-distribution problems.

For datasets such as Bias-in-Bios \citep{De-Arteaga-et-al-2019-bias}, a lot of existing works have simply relied on the scrubbed version provided in the dataset, however, we realize that the set of gender words scrubbed is very limited, hence, we extend this to include all gender words from \citep{kaneko-bollegala-2021-context} and gender names from \citep{hall-maudslay-etal-2019-name} to make the approach more effective.

\paragraph{Attribute swapping.}\label{para:Attribute swapping}
CDA has been explored and proposed as another bias mitigation strategy for downstream tasks, particularly in coreference resolution \citep{lu2020gender} and text classification \citep{park-etal-2018-reducing}. 
The idea is to generate counterfactuals by identifying attribute terms in the text instances and swapping them with equivalent terms belonging to the complementary group. 
We will term this approach \emph{Attribute swapping} to distinguish it from pretrained CDA in \autoref{para:pretrained_CDA}.
For pretrained CDA, the language model is (further) pretrained on a set of general corpora on Masked Language Modeling task, whereas in attribute swapping, counterfactuals are generated for each instance in the training data based on attribute terms; with the same $y$ labels assigned to both the original and counterfactual examples.
Whereas previous versions of CDA ignore names of people, which are major demographic cues,  \citet{hall-maudslay-etal-2019-name} propose a name intervention to generate counterfactuals for names as well. We use this adaptation for attribute swapping.

\subsection{Measuring extrinsic bias}
\label{ss:Evaluating extrinsic bias}
\paragraph{True positive rate difference (TPRD).}
We use TPRD used in related works \citep{De-Arteaga-et-al-2019-bias, prost-etal-2019-debiasing, jin-etal-2021-transferability, steed-etal-2022-upstream}. TPRD measures the gap in true positive rates between the predictions for two demographic groups: $TPRD = P(\hat{y}=1|y=1,A=a) - P(\hat{y}=1|y=1,A=a^{\prime})$.

\paragraph{Counterfactual Fairness.}
Since the test dataset follows the same biased distribution of the training set, we also measure the disparity in performance for an individual test instance if the gender attributes are swapped.
We measure counterfactual fairness \citep{kusner2017counterfactual} as the difference in the probability of a positive prediction between a test example and its counterfactual, following
$CF =\text{mean}(P(|x(\hat{Y}_{A\leftarrow a}=y|A=a,X=x) - P(\hat{Y}_{A\leftarrow a^\prime}=y|A=a,X=x)|))$.



\section{Probing for gender information}
\label{sec:probe}

We design a probe to investigate how much gender information is retained after \emph{mitigating} bias in pretrained language models.
Our goal is to measure the amount of gender information in the resulting embeddings of language models after mitigation techniques from \autoref{ss:mitigating-intrinsic-bias} have been applied to them.

\paragraph{Defining attribute and stereotype terms.}
We first identify two sets of attribute terms that define females and males respectively.
Let $W_f=\{she, female, woman, ...\}$ and $W_m=\{he, male, man, ...\}$ be the respective sets of female and male gender-defining words. The set $W=W_f \cup W_m$ is the union set of gender-defining words for both females and males.
Secondly, we define two sets of stereotype terms which are composed of target words associated with females and males respectively\footnote{We use both attribute and stereotype wordlists from \cite{kaneko-bollegala-2021-context}}.
Let $X_f=\{cheerleader, nurse, softball, ...\}$ and $X_m=\{warrior, baseball, engineer, ...\}$ be the respective sets of female and male stereotype words such that $X=X_f \cup X_m$.

\paragraph{Extracting word embeddings.}
We extract all sentences from a  text corpus containing words in each set of words.
Especially for stereotype terms, we ensure that the sentences do not also include words in the attribute wordlist since attribute words in the same context can introduce gender information in the embedding produced.
Let $S_w=\{s_{w_1},..., s_{w_n}\}$ be the set of all sentences containing a word $w$. We use the pretrained language model to extract embeddings for $w$ such that $w \rightarrow  E_w = \{e_{w_1}, ..., e_{w_n} \}$, where $e_{w_i} \in \mathbb{R}^n$ is the embedding for token $w$ in sentence $s_{w_i}$.
If $w$ appears in $s_{w_i}$ multiple times, then $|E_w| > n$.

\paragraph{Training the classifier.}
We split $W$ into a train and test sets $W_T$ and $W_I$ respectively such that $W_T \cap W_I = \emptyset$. 
This is to verify that the trained model is not merely learning to identify words that appear in the train set rather than words containing gender information in general.
We reserve the set of stereotypes $X$ solely for evaluating bias (inference) in the language models.

Taking $E_f=\{e_{f1}, e_{f2}, ...\}$ and $E_m=\{e_{m1}, e_{m2}, ...\}$ to be the set of all embeddings for female and male words respectively, given $E = {E_f \cup E_m}$, we train a classifier $\mathcal{C}:E\rightarrow\{f,m\}$ to predict whether an embedding $e$ belongs to $E_f$ or $E_m$.

\paragraph{Evaluating bias.}
To determine the conformity of a language model $\mathcal{L}$ to gender stereotypes $X$, we introduce two bias criteria (or notions of bias):

\textbf{Bias accuracy:} We compute the accuracy of $\mathcal{C}$ to correctly predict the gender association of all stereotype words in $X$ based on the embeddings produced by $\mathcal{L}$. A high accuracy depicts the presence of pro-stereotype information in the embeddings produced by the language model $\mathcal{L}$, even though the predictions could be of low confidence i.e. just enough gender information to correctly predict the gender association.

\textbf{Mean bias confidence:} We also consider the mean of the individual softmax scores for each prediction.
In the case of binary gender, a softmax value of 0.5 means the embedding is perfectly neutral, hence, we define mean bias confidence as
$\frac{1}{N}\sum_i^N |bias(e_i)-0.5|$. High mean confidence depicts an enormous amount of gender information contained in the embedding, although this information may not necessarily be pro-stereotype.

\paragraph{Randomization test.}
To test the efficacy of our method,
We carry out a randomization test by iteratively splitting $X = X_f \cup X_m$ into 
two random groups $X_{f^\prime}$ and $X_{m^\prime}$.
We use $X^{\prime(i)} = \{X_{f^\prime}^{(i)}$, $X_{m^\prime}^{(i)}$\} to define the random pair generated in the $i$th iteration.
We repeat this randomized split over 100 iterations; $X^{\prime(i)}\in\{X^{\prime(1)}, ..., X^{\prime(100)}\}$.
By computing a p-value based on a one-sample t-test, we compute if the mean of accuracy scores from the random samples $X^\prime$ is significantly different from the score from$\{X_f, X_m\}$.

\section{Investigating intrinsic bias in mitigated models with existing metrics}

We first use the two existing metrics defined in \autoref{ss:measuring-intrinsic-bias} (SEAT and LPBS) to quantify bias in BERT-large and ALBERT-large pretrained models. Using the same metrics, we then explore how intrinsic bias changes in mitigated versions of these models using the mitigation techniques outlined in \autoref{ss:mitigating-intrinsic-bias}: Context-debias, Sent-debias, and CDA pretraining.
This is important to evaluate the efficacy of these mitigation strategies, and to make relevant postulations.

\begin{figure}[tbh]
     \centering
          
    \label{fig:Intrinsic bias}
     \begin{subfigure}{0.45\textwidth}
         \centering
         \includegraphics[width=\textwidth]{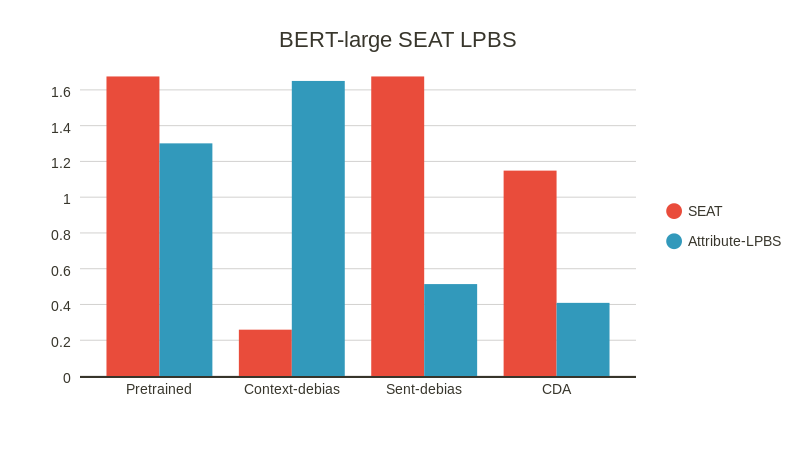}
         \caption{BERT-large model}
         \label{fig:Bert Large intrinstic bias}
     \end{subfigure}
\caption{Intrinsic bias scores using SEAT and LPBS. Results show inconsistencies in measuring bias between SEAT and LPBS for various mitigation strategies --- lower scores are desirable in both cases.}\label{fig:BERT intrinsic bias}
\end{figure}

\subsection{RQ1: Do intrinsic bias measures respond differently to bias mitigation techniques}\label{ss:rq1}
In \autoref{fig:BERT intrinsic bias}, we plot the intrinsic bias results\footnote{see \autoref{tab:intrinsic gender_info} and \autoref{fig:intrinsic bias} for full results.} of SEAT and LPBS on BERT-large.
The first observation we point out, particularly for Context-debias and Sent-debias, is the contradicting nature of bias scores obtained from SEAT and LPBS.
Whereas SEAT shows a drastic reduction in intrinsic bias for Context-debias, LPBS indicates worsening bias. The converse is true for Sent-debias where LPBS shows a significant reduction in intrinsic bias but SEAT shows worse scores.
This confirms our suspicion that different mitigation strategies respond differently to different metrics.

\begin{figure}[tb]
     \centering
     \begin{subfigure}{0.45\textwidth}
         \centering
         \includegraphics[width=\textwidth]{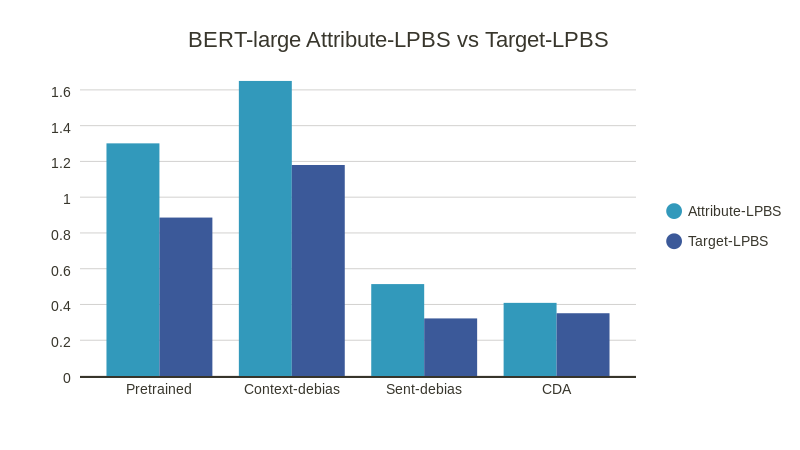}
         \label{fig:Bert LPBS Large intrinstic bias}
     \end{subfigure}
\caption{We compare LPBS in its default form (attribute-LPBS) and target-LPBS for BERT-large. The plot shows a general positive correlation between the two metrics. In both cases, Sent-debias maintains the lowest bias score --- lower scores are desirable in both cases. See \autoref{fig:intrinsic bias LPBS} for ALBERT-large.} \label{fig:BERT intrinsic bias LPBS}
\end{figure}

\paragraph{The use of some metrics can be problematic for evaluating some mitigation techniques for intrinsic bias.}
We find that some metrics are not ideal to use with some mitigation techniques due to differences in how various mitigation techniques interact with a language model.
Consider a metric like LPBS which measures the association between a stereotype term and a gender 
$\frac{P(A=a|X=x)}{P(A=a)}$.
This association is captured at the inner layers of the model, 
but since post-hoc approaches like Sent-debias do not alter the internal representation of the model, LPBS in its original form becomes ineffective --- \autoref{fig:LPBS_sent_debias}. 
We refer to this original form of LPBS as \textbf{\emph{attribute-LPBS}}.
Sent-debias ends up debiasing the attribute terms rather than the stereotype terms, thus, constantly producing low probability difference scores.
Even so, we find LPBS being used in this form with Sent-debias in works such as \citep{steed-etal-2022-upstream}. 
Using stereotype terms instead of attribute terms ${P(X=x|A=a)}$, which we refer to as \textbf{\emph{target-LPBS}},
is an option to solve this discrepancy, but given that many stereotype terms are usually out of vocabulary, these models will resort to wordpiece tokenization \cite{sennrich-etal-2016-neural} which will be more challenging to handle.
Nonetheless, in \autoref{fig:BERT intrinsic bias LPBS}, when the two versions of LPBS are compared, Sent-debias continues to have the best intrinsic fairness scores in both cases \footnote{For this comparison we only selected single-wordpiece attribute words to avoid challenges with multiple wordpieces.}.

\section{Probing further to uncover bias}
We use the methodology in \autoref{sec:probe} to probe for bias in pretrained language models and their bias-mitigated versions to see if the intrinsic bias mitigation techniques actually mitigate biases or superficially conceal them from commonplace metrics like SEAT and LPBS.
\citet{Gonen2019LipstickOA} discovered that, for non-contextualized word embeddings, the bias mitigation techniques proposed by \citep{bolukbasi2016man} were hiding bias instead of resolving bias. 
We use our probe to see if this is equally the case for intrinsic bias mitigation in contextualized embeddings.
The consequence of mitigation techniques superficially hiding bias is that downstream classifiers can learn to pick up residual traces of gender information.

We validate our results by carrying out a randomization test, as described in \autoref{sec:probe}, to test for statistical significance of our results.
We conduct the test under the null hypothesis $H_0$ that the prediction accuracy of our original gender-stereotype sets will not differ significantly from the random split. 
We show in \autoref{tab:detecting gender_info} the very low p-values that indicate statistical significant results, thus refuting $H_0$. 

\subsection{RQ2: Intrinsic bias techniques and metrics can hide bias instead of resolving it}\label{ss:rq2}

\begin{figure}[tbh]
     \centering
     \begin{subfigure}{0.45\textwidth}
         \centering
         \includegraphics[width=\textwidth]{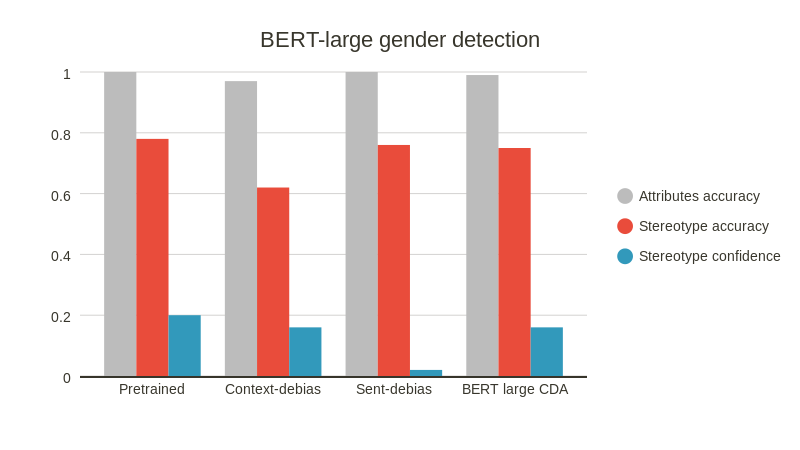}
         \label{fig:Bert Large gender detection}
     \end{subfigure}
     \caption{Gray bars indicate the accuracy of detecting gender information in attribute terms (high scores are desirable). Red bars indicate the accuracy of detecting gender information in stereotype terms (this should ideally be 0.5 in a fair classifier, showing the inability to correctly predict the gender association of stereotype terms). Blue bars indicate the average confidence of prediction: $\frac{1}{N}\sum_i^N |bias(e_i)-0.5|$ (low scores are desirable).}
     \label{fig:BERT Gender_detection}
\end{figure}

In \autoref{fig:BERT Gender_detection}, we show the results\footnote{see  \autoref{tab:detecting gender_info} and \autoref{fig:Gender_detection} for full results.} of our probe for BERT-large. We highlight some key observations. 

\paragraph{Bias accuracy and mean bias confidence do not correlate.}
We see from both plots in \autoref{fig:BERT Gender_detection} that, counter-intuitively, high bias accuracy does not concur with high mean bias confidence. Actually, Sent-debias, which consistently has the highest bias accuracy among the mitigation techniques, consistently has the lowest mean bias confidence. This indicates that these two notions of bias may not correspond. This, coupled with the finding in \autoref{fig:BERT intrinsic bias} about the inconsistency between SEAT and LPBS, suggests that intrinsic bias could be perceived or measured from conflicting perspectives.

\paragraph{Some mitigation techniques can conceal gender information.}
For Sent-debias in ALBERT-large, we can, almost to the same degree of accuracy as for the unmitigated pretrained model, correctly predict the association of stereotypes to their respective genders using a simple linear classifier even with a low confidence. This will render a mitigation technique useless if a downstream classifier can correctly predict gender.
This could only indicate that although models like Sent-debias significantly reduce the degree of association between stereotypes and their respective genders, hence the low confidence in prediction,  this reduction does not inhibit the capability to correctly predict the associated gender of stereotypes;
confirming that an intrinsic mitigation process can be rendered useless for downstream fairness even if the degree of association (confidence) is significantly reduced, so long as the information retained remains just enough to correctly associate the gender groups with their associated stereotypes.

\subsection{RQ3: Do intrinsic bias mitigation techniques in language models improve fairness in downstream tasks?}\label{ss:rq3}
To investigate the relationship between intrinsic and extrinsic gender bias, we train both BERT-large and ALBERT-large together with their intrinsic-bias-mitigated versions on the Bias-in-bios and Jigsaw datasets using a 10-fold train-test scheme to correctly predict the profession of each profile. We evaluate extrinsic bias based on gender TPRD and gender CF scores described in \autoref{ss:Evaluating extrinsic bias}.
We conduct the test under three settings: 1) Using the original version of the training data. 2) Using our attribute scrubbed version \autoref{para:Attribute scrubbing}. 3) Using the attribute-swap augmented version of the data \autoref{para:Attribute swapping}.
These three settings of the dataset are important for us to understand how intrinsic bias affects downstream fairness under different data settings.

\begin{figure*}
     \centering
     \begin{subfigure}[b]{0.45\textwidth}
         \centering
         \includegraphics[width=\textwidth]{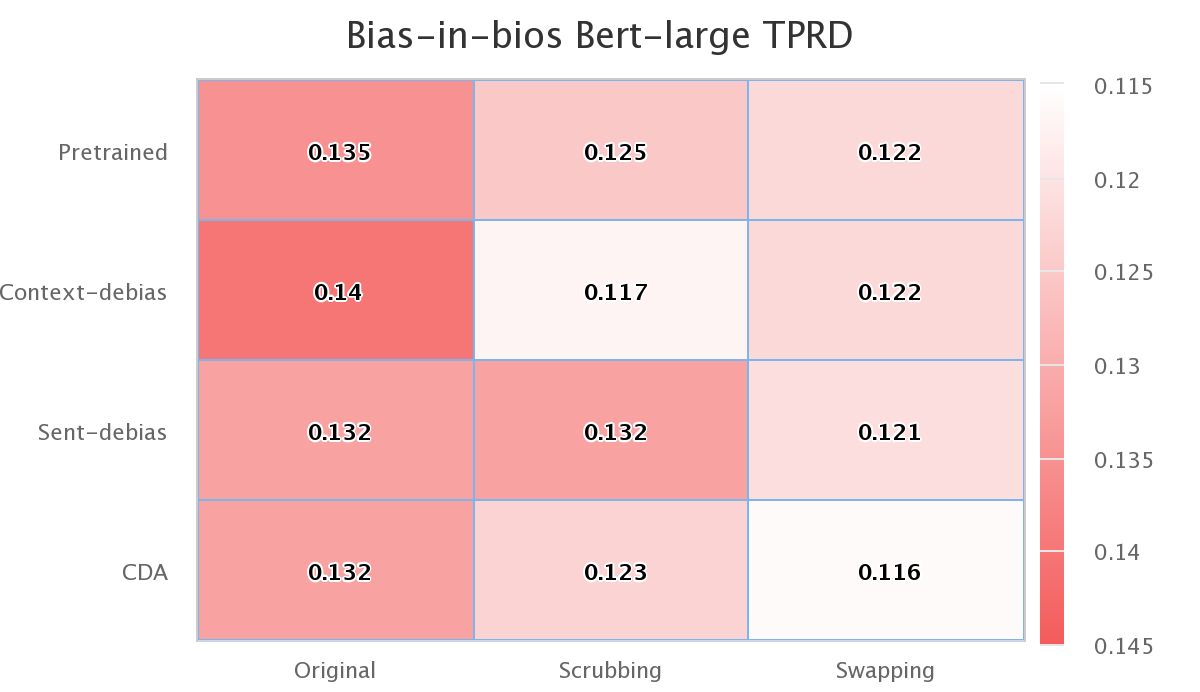}
         \label{fig:Bert Large Intrinsic-extrinsic TPRD-2}
     \end{subfigure}
     \hfill
     \begin{subfigure}[b]{0.45\textwidth}
         \centering
         \includegraphics[width=\textwidth]{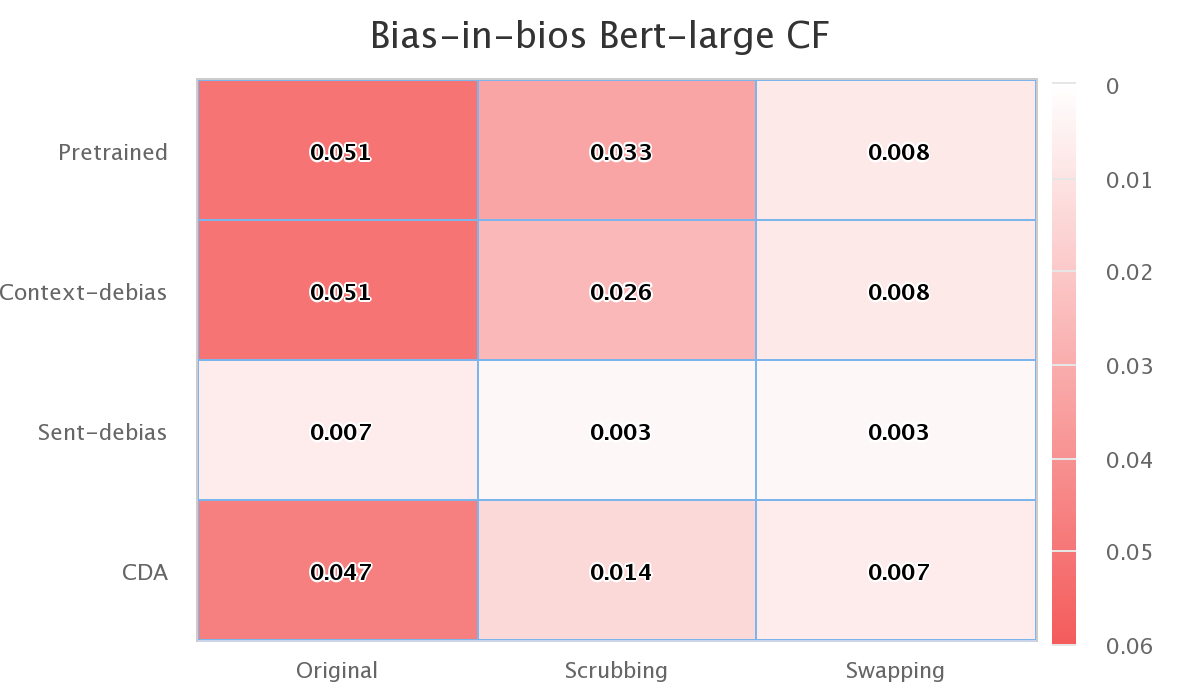}
         \label{fig:BERT Large Intrinsic-extrinsic CF}
     \end{subfigure}
     \caption{How intrinsic mitigation and downstream data intervention interact to influence fairness on the bias-in-bios data --- BERT-large.} \label{fig:bias-in-bios BERT Intrinsic-extrinsic bias}
\end{figure*}

In \autoref{fig:bias-in-bios BERT Intrinsic-extrinsic bias}, we show the extrinsic bias score \footnote{Full results with error margins are given in \autoref{tab:bias in bios Bert},  \autoref{tab:bias in bios Albert}, \autoref{tab:Jigsaw Bert} and \autoref{tab:Jigsaw Albert}.} for BERT-large and its \emph{debiased} versions, and the corresponding downstream data type for the Bias-in-Bios dataset --- see \autoref{fig:bias-in-bios Intrinsic-extrinsic bias} and \autoref{fig:jigsaw extrinsic-Intrinsic bias}  for all models and datasets.

\paragraph{Intrinsic mitigation techniques do not show a significant improvement on TPRD.}
We observe that all the intrinsic bias mitigation interventions we consider do not significantly improve bias in TPRD on their own.
There is only a slight improvement or worsening bias in some cases. For Context-debias, we postulate this could be due to the limited list of words used to mitigate bias. If these limited words are not key terms in the downstream task or do not exist in the downstream text instances, the mitigation may not have a consequential effect on the downstream task. 

\paragraph{Downstream data processing significantly improves downstream fairness.}
Secondly, we realize that the downstream data has a lot more material effect on downstream fairness. The data processing techniques: \emph{attribute scrubbing and attribute swapping} showed improvements in TPRD producing the best results.

\paragraph{Combining intrinsic CDA and Downstream CDA produces best TPRD results.}
Thirdly, we realize that although the downstream data seems key to downstream fairness, a combination of intrinsic and downstream data intervention produces even better extrinsic fairness results.

\paragraph{Sent-debias significantly improves CF scores.}
Sent-debias is the only intrinsic technique that shows significant improvement in counterfactual fairness. Context-debias and CDA produce worse CF scores in some cases. 
We observe that the mode of application of Sent-debias in downstream classification tasks does not entirely align with the notion of intrinsic since gender information is removed from the sentence-level representation instead of word-level representation. 

\paragraph{Combining Sent-Debias and fair downstream data produces the best CF scores.}
We also realize from BERT (\autoref{fig:bias-in-bios BERT Intrinsic-extrinsic bias}) and ALBERT that combining Sent-debias as a mitigation technique and a downstream data intervention produces the best CF scores.

\begin{figure}[tbh]
     \centering
     \begin{subfigure}[b]{0.48\textwidth}
         \centering
         \includegraphics[width=\textwidth]{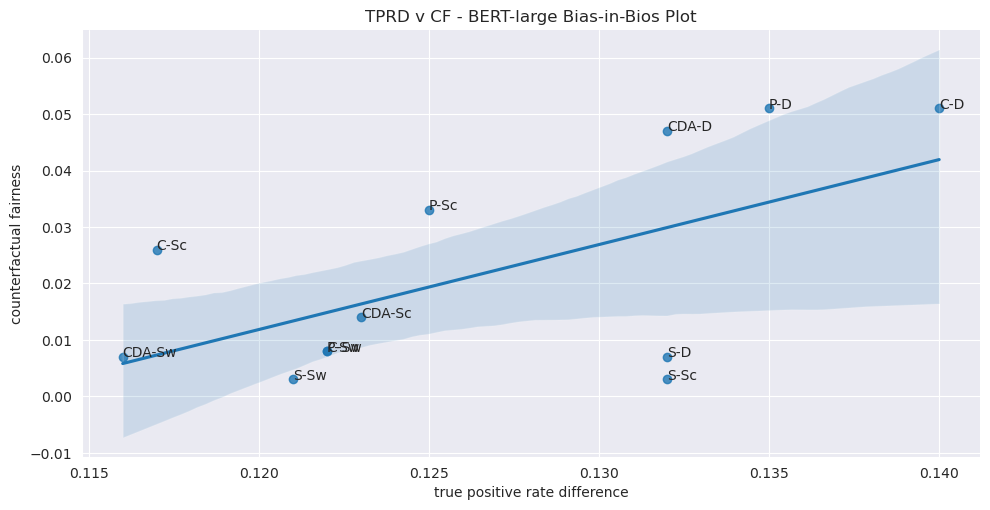}
         \label{fig:TPR-CF-Bert-Large}
     \end{subfigure}
     \caption{Relationship between TPRD and CF based on results from BERT model and Bias-in-Bios dataset. See \autoref{fig:TPRD-CF} for ALBERT-large.
}
    \label{fig:BERT TPRD-CF}
\end{figure}

\paragraph{TPRD vs Counterfactual-fairness.}
Here,  we look at the relationship between TPRD and counterfactual fairness. As both measures are used as extrinsic fairness metrics, we aim to see how both results correlate. 
From \autoref{fig:BERT TPRD-CF} we find a positive correlation in general between the two metrics.
The downstream data interventions tend to improve both TPRD and CF the most. 

\section{Related work}
\citet{park-etal-2018-reducing} study the effects of three techniques to mitigate gender bias in abusive language detection; \emph{debiased word embeddings} (based on the approach by \citet{bolukbasi2016man}), \emph{gender swap} (counterfactual augmentation of classification data), \emph{bias finetuning} using non-contextualized word embeddings: word2vec \citep{mikolov2013distributed} and FastText \citep{bojanowski2017enriching}.
They found that a combination of debiasing and gender swapping produced the best fairness results when implemented with a Gated Recurrent Unit (GRU).

\citet{prost-etal-2019-debiasing} examine three debiasing techniques for mitigating gender bias in downstream text classification using bias-in-bios \citep{de2019bias}, with non-contextualized embeddings.
Specifically, they examine \emph{scrubbing} (deletion of identity terms) and approaches that focus on non-contextualized embeddings namely: \emph{debiasing} (based on neutralization and equalization by \citet{bolukbasi2016man}), and \emph{strong debiasing} \citet{prost-etal-2019-debiasing}. 
Experimenting with Glove embeddings \citep{pennington2014glove}, they show that strong debiased produces the best fairness results whilst maintaining a good accuracy score (second only to the unmitigated model). 
They again show that standard debiasing \citep{bolukbasi2016man} can be counter-productive in terms of fairness; it can rather reduce fairness in downstream text classification.

\citet{steed-etal-2022-upstream} meticulously investigate the effect of intrinsic bias in downstream text classification in RoBERTa. They also conclude that intrinsic/upstream bias does not significantly contribute to downstream bias. 
However, as we have previously discussed, using the default form of Log probability Bias score (attribute-LPBS) \citep{kurita-etal-2019-measuring} which they adopt to measure intrinsic bias may not be an effective technique to use with Sent-debias.

\citet{orgad-etal-2022-gender}, the closest and concurrent with our work, investigate the relationship between intrinsic and extrinsic bias and also develop a probe based on a classification-based technique \emph{minimum description length} \citep{voita-titov-2020-information} to detect bias in RoBERTa. They reach similar conclusions indicating how intrinsic metrics, like CEAT in their case, could conceal bias.

\section{Conclusion}
In this paper, we develop a probe to investigate intrinsic bias in two language models, BERT-large and ALBERT-large on two text classification tasks. We use this probe to answer key research questions outlined for this research.
We find that intrinsic bias metrics can be sensitive to certain intrinsic bias mitigation techniques.
We also show that intrinsic bias mitigation techniques and metrics are capable of concealing bias instead of resolving it.
We discover that intrinsic bias mitigation techniques we considered do not significantly improve fairness in text classification when used without other fairness interventions like data pre-processing.
We recommend that intrinsic bias mitigation should ideally be combined with other fairness interventions.

\section*{Limitations}
One obvious limitation of this work is the restriction to the use of English language as a basis. In future, we will explore how these conclusions vary in other language settings.
Secondly, although we aimed to make the list of gender and attribute terms used in our experiments as extensive as possible, it is nearly impossible to cover all possible gender and target words in such contexts. Nonetheless, for the scope under consideration, we believe our compilation is extensive enough to give a general outlook and to draw the necessary conclusions.
Finally, our work is heavily based on binary notions of gender which is a limitation considering the growing use of non-binary categorizations of gender and other biases which we outline in the ethical considerations section.


\section*{Ethical considerations}
\label{sec:ethical considerations}

For practical reasons such as access to datasets and resources on gender bias, we limit our work to a binary representation of gender. We draw readers' attention to the fact that non-binary gender representations are nuanced and intricate \citep{dev-etal-2021-harms}, as such, this should be in cognizance when applying conclusions from this work in non-binary settings. Nonetheless, considering binary gender as a base form of gender categorization, the insights and conclusions from this work can form the baseline for exploring more complex gender categorizations.




\section*{Acknowledgements}
Ewoenam Kwaku Tokpo and Pieter Delobelle received funding from the Flemish Government under the ``Onderzoeksprogramma Artificiële Intelligentie (AI) Vlaanderen'' programme.
Pieter Delobelle was also supported by the Research Foundation - Flanders (FWO) under EOS No. 30992574 (VeriLearn).
Bettina Berendt received funding from the German Federal Ministry of Education and Research (BMBF) – Nr. 16DII113.

\bibliographystyle{acl_natbib}
\bibliography{eacl2023}

\appendix

\section*{Appendix}
\label{sec:appendix}


\section{Word lists for experiments}
We generate the wordlist we use, given below, from the more extensive wordlist from \cite{kaneko-bollegala-2021-context}.
To mitigate the effect of multi-wordpiece tokens, we only select tokens with a single wordpiece. Hence we obtain the following wordlists:

\subsection{attribute terms}\label{ss:attribute terms}
\begin{itemize}
    \item \textbf{female list:} 'witches', 'mothers', 'diva', 'actress', 'mama', 'dowry', 'princess', 'abbess', 'women', 'widow', 'ladies', 'madam', 'baroness', 'niece', 'lady', 'sister', 'nun', 'her', 'mare', 'convent', 'ladies', 'queen', 'maid', 'chick', 'empress', 'mommy', 'feminism', 'gal', 'estrogen', 'goddess', 'aunt', 'hostess', 'wife', 'mom', 'females', 'ma', 'belle', 'maiden', 'witch', 'miss', 'cow', 'granddaughter', 'her', 'mistress', 'nun', 'actresses', 'girlfriend', 'lady', 'maternal', 'ladies', 'sorority', 'duchess', 'ballerina', 'fiancee', 'wives', 'maternity', 'she', 'heroine', 'queens', 'sisters', 'stepmother', 'daughter', 'lady', 'daughters', 'mistress', 'hostess', 'nuns', 'priestess', 'filly', 'herself', 'girls', 'lady', 'vagina', 'wife', 'mother', 'female', 'womb', 'heiress', 'waitress', 'woman', 'bride', 'grandma', 'bride', 'gal', 'lesbian', 'ladies', 'girl', 'grandmother', 'mare', 'maternity', 'nuns'
    
    \item \textbf{male list:} 'wizards', 'fathers', 'actor', 'bachelor', 'papa', 'dukes', 'hosts', 'airmen', 'penis', 'prince', 'governors', 'abbot', 'men', 'gentlemen', 'sir', 'baron', 'gods', 'nephew', 'lord', 'brother', 'priest', 'his', 'marquis', 'princes', 'emperors', 'stallion', 'chairman', 'monastery', 'priests', 'king', 'spokesman', 'tailor', 'cowboys', 'dude', 'emperor', 'daddy', 'guys', 'guy', 'godfather', 'god', 'patriarch', 'uncle', 'sir', 'brotherhood', 'host', 'testosterone', 'husband', 'dad', 'steward', 'males', 'pa', 'beau', 'stud', 'bachelor', 'wizard', 'sir', 'bull', 'grandson', 'horsemen', 'rooster', 'bachelor', 'him', 'master', 'lad', 'policeman', 'monk', 'actors', 'boyfriend', 'statesman', 'paternal', 'landlord', 'brethren', 'lords', 'fraternity', 'duke', 'fiance', 'colts', 'husbands', 'he', 'businessman', 'hero', 'deer', 'kings', 'brothers', 'masters', 'stepfather', 'son', 'cowboy', 'sons', 'baritone', 'salesman', 'monks', 'lads', 'congressman', 'priest', 'barons', 'beard', 'colt', 'himself', 'boys', 'lions', 'gentleman', 'penis', 'his', 'bulls', 'uncles', 'lion', 'sorcerer', 'father', 'male', 'sperm', 'prostate', 'businessmen', 'heir', 'waiter', 'headmaster', 'man', 'governor', 'god', 'grandpa', 'groom', 'dude', 'gay', 'boy', 'grandfather', 'priests'
\end{itemize}

\subsection{List of stereotypes(target words)}\label{ss:target-words}
\begin{itemize}
    \item \textbf{female list:} {'beauty', 'blonde', 'cheerleading', 'cooking', 'flirt', 'flower', 'gossip', 'housekeeper', 'jewelry', 'nanny', 'nurse', 'nurses', 'pink', 'pretty', 'receptionist', 'sewing', 'softball', 'makeup', 'therapist', 'maid', 'chairperson'}
    
    \item \textbf{male list:} {'warrior', 'strong', 'terrorist', 'soldier', 'sniper', 'baseball', 'sergeant', 'bodyguard', 'boss', 'boxer', 'captain', 'carpenter', 'chancellor', 'colonel', 'commander', 'conductor', 'diplomat', 'drummer', 'engineer', 'gangster', 'geek', 'guitarist', 'industrialist', 'marshal', 'mechanic', 'philosopher', 'physicist', 'scientist', 'rapper', 'mechanic', 'carpenter', 'clergy'}
\end{itemize}

\section{Intrinsic metrics}
\subsection{Context-debias} 
\label{ss:Context-debias formula}
\[
L_i = \sum_{t\in \mathcal{V}_t} \sum_{x \in \Omega(t)} \sum_{a \in \mathcal{V}_a} (v_i(a)^TE_I(t,x;\theta_e))^2))
\]
\[
L_{reg} = \sum_{x \in A} \sum_{w \in x} \sum_{i=1} ||E_i(w,x;\theta_e) - E_i(w,x;\theta_pre)||^2 
\]
\[
L = \alpha L_i + \beta L_{reg}
\]
Where $L_i$ is the orthogonality constraint such that $E_i(w, x; \theta_e )$ denotes the
embedding of token $w$ in the $i$-th layer of a contextualised word embedding model $E$ with parameters $\theta_e$. $v_i(a)$ is the non-contextualised embedding of an attribute word $a$.
$L_{reg}$ is a regularizer that constrains the Euclidean distance between the contextualized word embedding of a word $w$ in the $i$th layer in the original pretrained model with parameters $\theta_{pre}$.

\subsection{Sent-debias}
\label{ss:Sent-debias formula}

\[
v = PCA_k(\cup_{j=1} \cup_{w\in\mathcal{R}_j} (w-\mu_j))
\]
\[
h_v = \sum_{j=1}^k \langle h,v_j \rangle v_j
\]
\[
\hat{h} = h - h_v
\]
where $v$ represents the top-$k$ gender subspace, $h$ the representation of a given embedding, $h_v$ the projection of $h$ onto $v$, and $\hat{h}$ the resulting debiased subspace.

\section{Experiment set-up}
We train all models for 4 epochs, a learning rate of $2e^{-5}$ on Tesla V100-SXM3-32GB. We use a sequence length of 100 as the default for all extrinsic tasks.
For all downstream tasks, we use a one-layer linear classifier.

\section{Datasets}
\label{sec:Datasets}
\paragraph{Bias-in-bios} \citep{De-Arteaga-et-al-2019-bias} consists of online English biographies of people. We use this dataset to predict the occupations of people. Since the gender labels are provided, we can compute disparities in predictive performance between male and female gender groups.

We select the top 7 female and top 7 male professions based on the gender percentages. We take these two sets to represent female and male dominated jobs respectively.

\paragraph{Jigsaw dataset}
[\emph{https://www.kaggle.com/c/jigsaw-unintended-bias-in-toxicity-classification/data}] form the online platform \emph{Civil comments} contains online comments. This dataset is scored from 0 to 10 with the perceived gender of the targets of these comments as well as the toxicity levels of these comments. We use this dataset to predict whether or not a comment is toxic against the target persons. 
We use gender labels to compute predictive disparities.

We select all comments with toxicity score above 0.5 as the toxic comments and those with a score of exactly 0 as the non-toxic ones in order to eliminate fine-margins. 
We use the same technique to annotate gender by labeling a gender positive if it scores above 0.5 whilst its complimentary gender has a score of exactly 0. 

\section{Measuring intrinsic bias with LPBS when debiasing is done with Sent-debias}
\begin{figure}[tbh]
    \centering
    \includegraphics[height=120pt, width=0.3 \textwidth]{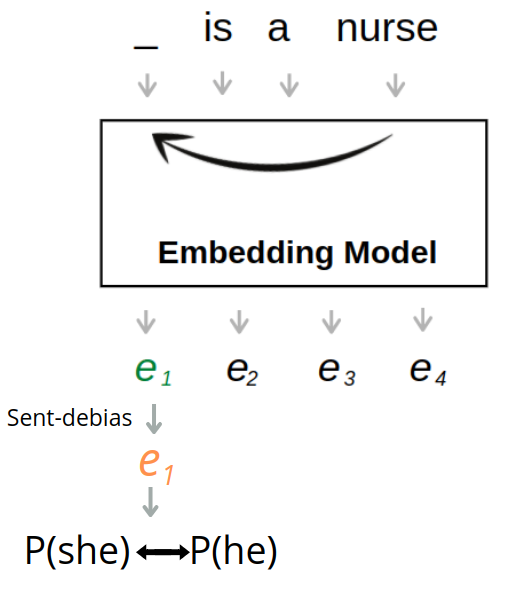}
    \caption{Measuring intrinsic bias in Sent-debias with target-LPBS.}
    \label{fig:LPBS_sent_debias}
\end{figure}

\section{Results}
\label{sec:results}








     

\begin{table*}[tbh]
\caption{Intrinsic bias scores from SEAT and LPBS. For SEAT we capture the \emph{test statistic score}: \textbf{SEAT test score} and the \emph{effect size}: \textbf{SEAT eff. size}. For LPBS we capture results from two variants: \emph{attribute-LPBS}: \textbf{attr-LPBS} and \emph{target-LPBS}: \textbf{target-LPBS}.}
\label{tab:intrinsic gender_info}
    \begin{subtable}[c]{0.5\textwidth}
    \centering
        \caption{Intrinsic bias BERT-large-uncased}\label{tab:detecting gender_info Bert_large}
\resizebox{\linewidth}{!}
{
\renewcommand{\arraystretch}{1} 
  \begin{tabular}{l|l|l|l|l}
    \toprule
     \textbf{Model}& \textbf{SEAT test score}& \textbf{SEAT eff. size}&\textbf{attr-LPBS}&\textbf{target-LPBS}\\ \hline
        \midrule
    pretrained  & 0.055& 1.675  & 1.301 & 0.886
\\
    Context-debias  &0.002 & 0.260  & 1.650 & 1.180

\\ \midrule
    Sent-debias &0.055 & 1.675 & 0.514 & 0.322

\\
    CDA & 0.033& 1.148 & 0.409 & 0.351

\\ \bottomrule
  \end{tabular}
}

    \end{subtable}
    \hfill
    \begin{subtable}[c]{0.5\textwidth}
    \centering
        \caption{Intrinsic bias ALBERT-large }\label{tab:detecting gender_info Albert_large}
\centering
\resizebox{\linewidth}{!}
{%
\renewcommand{\arraystretch}{1} 
  \begin{tabular}{l|l|l|l|l}
    \toprule
     \textbf{Model}& \textbf{SEAT test score}& \textbf{SEAT eff. size}&\textbf{attr-LPBS}&\textbf{target-LPBS}\\ \hline
        \midrule
    pretrained &0.051 & 1.715  & 0.870 & 0.446

\\
    Context-debias  & 0.003& 0.244 & 1.343 & 0.346

\\ \midrule
    Sent-debias &0.051 & 1.717  & 0.187 & 0.111

\\
    CDA & 0.006& 1.170 & 0.654 & 0.772
\\ \bottomrule
  \end{tabular}
}
     \end{subtable}
     
     \label{tab:temps}
\end{table*}

\begin{table*}[tbh]
\caption{Results of gender detection probe: \textbf{Gender acc} denotes the accuracy of detecting gender information in attribute/gender terms (high scores are desirable). \textbf{Stereotype acc} indicates the accuracy of detecting gender information in stereotype terms (this should ideally be 0.5 in a fair classifier, showing the inability to correctly predict the gender association of stereotype terms). \textbf{Stereotype conf} indicate the average confidence of prediction of stereotypes: $\frac{1}{N}\sum_i^N |bias(e_i)-0.5|$ (low scores are desirable). P-values are from the randomization test.}\label{tab:intrinsic gender_info Bert_large}
\label{tab:detecting gender_info}
    \begin{subtable}[tbh]{0.49\textwidth}
        \caption{Detecting gender information BERT-large-uncased}
\centering
\resizebox{\linewidth}{!}{%
\renewcommand{\arraystretch}{1} 
  \begin{tabular}{l|l|l|l|l}
    \toprule
     \textbf{Model}& \textbf{Gender acc}&\textbf{Stereotype acc} &\textbf{Stereotype conf} & \textbf{p-value}\\ \hline
        \midrule
    pretrained  & 1.0 & 0.78 & 0.20& 9.055e-73\\
    Context-debias  &  {0.97}  & {0.62} & 0.16 & 1.398e-51
\\ \midrule
    Sent-debias & 1.0 & 0.76 & 0.02 & 8.772e-70
\\
    CDA & 0.99 & 0.75 &0.16 & 7.824e-77
\\ \bottomrule
  \end{tabular}
}
    \end{subtable}
    \hfill
    \begin{subtable}[tbh]{0.49\textwidth}
    \caption{Detecting gender information Alert-large-uncased}\label{tab:intrinsic gender_info Albert_large}
        \centering
\resizebox{\linewidth}{!}{%
\renewcommand{\arraystretch}{1} 
  \begin{tabular}{l|l|l|l|l}
    \toprule
     \textbf{Model}& \textbf{Gender acc}&\textbf{Stereotype acc} & \textbf{Stereotype conf} & \textbf{p-value}\\ \hline
        \midrule
    pretrained  & 1.0 & 0.83
 & 0.33 & 2.763e-71
\\
    Context-debias  &  {0.97}  & {0.63
} & 0.19 & 1.700e-48

\\ \midrule
    Sent-debias & 1.0 & 0.82
 &0.02 & 2.335e-95

\\
    CDA & 0.94 & 0.63
 & 0.12 & 3.497e-61

\\ \bottomrule
  \end{tabular}
}
     \end{subtable}
     \label{tab:temps-2}
\end{table*}


\begin{table*}[tbh]
\caption{Extrinsic bias scores from Bias in Bios - BERT-large: We capture true positive rate difference, false positive rate difference, counterfactual fairness, accuracy scroes for both gender groups, counterfactual augmented true positive rate difference and counterfactual augmented false positive rate diference scores.}\label{tab:bias in bios Bert}
\centering
\resizebox{\linewidth}{!}{%
\renewcommand{\arraystretch}{1} 
  \begin{tabular}{l|l|llll|l|llll}
    \toprule
      \hline
      \textbf{Model} & \textbf{Data} & \textbf{TPRD}  & \textbf{FPRD}& \textbf{ACC-F}&\textbf{ACC-M} & \textbf{CF} & \textbf{CF-TPRD} &
      \textbf{CF-FPRD} & \textbf{CF-ACC-F} & \textbf{CF-ACC-M} \\\hline
        \midrule
    Pretrained & Default  & 0.135$\pm$ 0.03   &  0.012$\pm$ 0.0  & 0.984$\pm$ 0.0 & 0.982$\pm$ 0.0 & 0.051$\pm$ 0.0 & 0.116$\pm$ 0.02 & 0.008$\pm$ 0.0 & 0.982$\pm$ 0.0 & 0.981$\pm$ 0.0 \\ 
    Pretrained & Scrubbing  & 0.125 $\pm$ 0.02 & 0.011 $\pm$ 0.0 & 0.984 $\pm$ 0.0 & 0.982 $\pm$ 0.0 & 0.033 $\pm$ 0.0 & 0.113 $\pm$ 0.02 & 0.008 $\pm$ 0.0 & 0.983 $\pm$ 0.0 & 0.981 $\pm$ 0.0 \\ 
    Pretrained & Swapping  & 0.122 $\pm$ 0.02 & 0.008 $\pm$ 0.0 & 0.983 $\pm$ 0.0 & 0.981 $\pm$ 0.0 & 0.008 $\pm$ 0.0 & 0.12 $\pm$ 0.02 & 0.008 $\pm$ 0.0 & 0.983 $\pm$ 0.0 & 0.981 $\pm$ 0.0  \\ \hline
    Context-debias & Default & 0.14 $\pm$ 0.03 & 0.012 $\pm$ 0.0 & 0.984 $\pm$ 0.0 & 0.982 $\pm$ 0.0 & 0.051 $\pm$ 0.0 & 0.119 $\pm$ 0.02 & 0.008 $\pm$ 0.0 & 0.982 $\pm$ 0.0 & 0.98 $\pm$ 0.0 \\
    Context-debias & Scrubbing & 0.117 $\pm$ 0.03 & 0.01 $\pm$ 0.0 & 0.984 $\pm$ 0.0 & 0.981 $\pm$ 0.0 & 0.026 $\pm$ 0.0 & 0.109 $\pm$ 0.03 & 0.009 $\pm$ 0.0 & 0.983 $\pm$ 0.0 & 0.981 $\pm$ 0.0 \\
    Context-debias & Swapping & 0.122 $\pm$ 0.03 & 0.009 $\pm$ 0.0 & 0.983 $\pm$ 0.0 & 0.981 $\pm$ 0.0 & 0.008 $\pm$ 0.0 & 0.12 $\pm$ 0.03 & 0.009 $\pm$ 0.0 & 0.983 $\pm$ 0.0 & 0.981 $\pm$ 0.0 \\ \hline
    CDA& Default& 0.132 $\pm$ 0.02 & 0.012 $\pm$ 0.0 & 0.983 $\pm$ 0.0 & 0.982 $\pm$ 0.0 & 0.047 $\pm$ 0.0 & 0.111 $\pm$ 0.02 & 0.008 $\pm$ 0.0 & 0.982 $\pm$ 0.0 & 0.98 $\pm$ 0.0 \\
    CDA & Scrubbing & 0.123 $\pm$ 0.03 & 0.009 $\pm$ 0.0 & 0.982 $\pm$ 0.0 & 0.981 $\pm$ 0.0 & 0.014 $\pm$ 0.0 & 0.121 $\pm$ 0.03 & 0.009 $\pm$ 0.0 & 0.982 $\pm$ 0.0 & 0.981 $\pm$ 0.0 \\ 
    CDA & Swapping & 0.116 $\pm$ 0.02 & 0.009 $\pm$ 0.0 & 0.983 $\pm$ 0.0 & 0.981 $\pm$ 0.0 & 0.007 $\pm$ 0.0 & 0.116 $\pm$ 0.02 & 0.009 $\pm$ 0.0 & 0.983 $\pm$ 0.0 & 0.981 $\pm$ 0.0  \\ \hline
    Sent-debias & Original & 0.132 $\pm$ 0.02 & 0.012 $\pm$ 0.0 & 0.983 $\pm$ 0.0 & 0.982 $\pm$ 0.0 & 0.007 $\pm$ 0.0 & 0.115 $\pm$ 0.02 & 0.008 $\pm$ 0.0 & 0.981 $\pm$ 0.0 & 0.98 $\pm$ 0.0 \\
    Sent-debias & Scrubbing & 0.132 $\pm$ 0.02 & 0.01 $\pm$ 0.0 & 0.983 $\pm$ 0.0 & 0.981 $\pm$ 0.0 & 0.003 $\pm$ 0.0 & 0.122 $\pm$ 0.03 & 0.008 $\pm$ 0.0 & 0.982 $\pm$ 0.0 & 0.98 $\pm$ 0.0 \\
    Sent-debias & Swapping & 0.121 $\pm$ 0.02 & 0.009 $\pm$ 0.0 & 0.982 $\pm$ 0.0 & 0.98 $\pm$ 0.0 & 0.003 $\pm$ 0.0 & 0.119 $\pm$ 0.02 & 0.009 $\pm$ 0.0 & 0.982 $\pm$ 0.0 & 0.98 $\pm$ 0.0 \\
  \end{tabular}
}
\end{table*}

\begin{table*}[tbh]
\caption{Extrinsic bias scores from Bias in Bios - ALBERT-large}\label{tab:bias in bios Albert}
\centering
\resizebox{\linewidth}{!}{%
\renewcommand{\arraystretch}{1} 
  \begin{tabular}{l|l|llll|l|llll}
    \toprule
      \hline
      \textbf{Model} & \textbf{Data} & \textbf{TPRD}  & \textbf{FPRD}& \textbf{ACC-F}&\textbf{ACC-M} & \textbf{CF} & \textbf{CF-TPRD} &
      \textbf{CF-FPRD} & \textbf{CF-ACC-F} & \textbf{CF-ACC-M} \\\hline
        \midrule
    Pretrained & Default & 0.138 $\pm$ 0.02 & 0.013 $\pm$ 0.0 & 0.982 $\pm$ 0.0 & 0.979 $\pm$ 0.0 & 0.06 $\pm$ 0.0 & 0.118 $\pm$ 0.02 & 0.009 $\pm$ 0.0 & 0.979 $\pm$ 0.0 & 0.978 $\pm$ 0.0 \\ 
    Pretrained & Scrubbing & 0.127 $\pm$ 0.01 & 0.01 $\pm$ 0.0 & 0.982 $\pm$ 0.0 & 0.978 $\pm$ 0.0 & 0.021 $\pm$ 0.0 & 0.12 $\pm$ 0.01 & 0.009 $\pm$ 0.0 & 0.981 $\pm$ 0.0 & 0.978 $\pm$ 0.0   \\ 
    Pretrained & Swapping & 0.121 $\pm$ 0.02 & 0.009 $\pm$ 0.0 & 0.981 $\pm$ 0.0 & 0.979 $\pm$ 0.0 & 0.01 $\pm$ 0.0 & 0.121 $\pm$ 0.02 & 0.009 $\pm$ 0.0 & 0.981 $\pm$ 0.0 & 0.979 $\pm$ 0.0 \\ \hline
    Context-debias & Default & 0.137 $\pm$ 0.02 & 0.012 $\pm$ 0.0 & 0.983 $\pm$ 0.0 & 0.98 $\pm$ 0.0 & 0.062 $\pm$ 0.0 & 0.114 $\pm$ 0.02 & 0.009 $\pm$ 0.0 & 0.981 $\pm$ 0.0 & 0.978 $\pm$ 0.0 \\
    Context-debias & Scrubbing & 0.129 $\pm$ 0.02 & 0.01 $\pm$ 0.0 & 0.98 $\pm$ 0.0 & 0.978 $\pm$ 0.0 & 0.021 $\pm$ 0.0 & 0.124 $\pm$ 0.02 & 0.009 $\pm$ 0.0 & 0.979 $\pm$ 0.0 & 0.977 $\pm$ 0.0 \\
    Context-debias & Swapping & 0.116 $\pm$ 0.02 & 0.009 $\pm$ 0.0 & nan $\pm$ nan & nan $\pm$ nan & 0.011 $\pm$ 0.0 & 0.115 $\pm$ 0.02 & 0.009 $\pm$ 0.0 & 0.981 $\pm$ 0.0 & 0.978 $\pm$ 0.0 \\ \hline
    CDA& Default & 0.142 $\pm$ 0.02 & 0.013 $\pm$ 0.0 & 0.981 $\pm$ 0.0 & 0.978 $\pm$ 0.0 & 0.063 $\pm$ 0.0 & 0.121 $\pm$ 0.02 & 0.009 $\pm$ 0.0 & 0.979 $\pm$ 0.0 & 0.977 $\pm$ 0.0 \\
    CDA & Scrubbing & 0.127 $\pm$ 0.02 & 0.01 $\pm$ 0.0 & 0.979 $\pm$ 0.0 & 0.977 $\pm$ 0.0 & 0.012 $\pm$ 0.0 & 0.125 $\pm$ 0.02 & 0.01 $\pm$ 0.0 & 0.979 $\pm$ 0.0 & 0.977 $\pm$ 0.0 \\ 
    CDA & Swapping & 0.115 $\pm$ 0.01 & 0.009 $\pm$ 0.0 & 0.98 $\pm$ 0.0 & 0.977 $\pm$ 0.0 & 0.011 $\pm$ 0.0 & 0.116 $\pm$ 0.02 & 0.009 $\pm$ 0.0 & 0.98 $\pm$ 0.0 & 0.977 $\pm$ 0.0 \\ \hline
    Sent-debias & Original & 0.145 $\pm$ 0.02 & 0.013 $\pm$ 0.0 & 0.981 $\pm$ 0.0 & 0.978 $\pm$ 0.0 & 0.007 $\pm$ 0.0 & 0.119 $\pm$ 0.03 & 0.009 $\pm$ 0.0 & 0.978 $\pm$ 0.0 & 0.976 $\pm$ 0.0\\
    Sent-debias & Scrubbing & 0.12 $\pm$ 0.02 & 0.01 $\pm$ 0.0 & 0.978 $\pm$ 0.0 & 0.976 $\pm$ 0.0 & 0.002 $\pm$ 0.0 & 0.108 $\pm$ 0.02 & 0.009 $\pm$ 0.0 & 0.978 $\pm$ 0.0 & 0.976 $\pm$ 0.0  \\
    Sent-debias & Swapping & 0.117 $\pm$ 0.03 & 0.009 $\pm$ 0.0 & 0.981 $\pm$ 0.0 & 0.978 $\pm$ 0.0 & 0.002 $\pm$ 0.0 & 0.114 $\pm$ 0.02 & 0.009 $\pm$ 0.0 & 0.981 $\pm$ 0.0 & 0.978 $\pm$ 0.0\\
  \end{tabular}
}
\end{table*}

\begin{table*}[tbh]
\caption{Extrinsic bias scores from Jigsaw - BERT-large}\label{tab:Jigsaw Bert}
\centering
\resizebox{\linewidth}{!}{%
\renewcommand{\arraystretch}{1} 
  \begin{tabular}{l|l|llll|l|llll}
    \toprule
      \hline
      \textbf{Model} & \textbf{Data} & \textbf{TPRD}  & \textbf{FPRD}& \textbf{ACC-F}&\textbf{ACC-M} & \textbf{CF} & \textbf{CF-TPRD} &
      \textbf{CF-FPRD} & \textbf{CF-ACC-F} & \textbf{CF-ACC-M} \\\hline
        \midrule
    Pretrained & Default & 0.044 $\pm$ 0.02 & 0.118 $\pm$ 0.04 & 0.896 $\pm$ 0.01 & 0.934 $\pm$ 0.0 & 0.116 $\pm$ 0.01 & 0.061 $\pm$ 0.01 & 0.029 $\pm$ 0.02 & 0.866 $\pm$ 0.01 & 0.902 $\pm$ 0.01 \\ 
    Pretrained & Scrubbing & 0.026 $\pm$ 0.02 & 0.076 $\pm$ 0.03 & 0.891 $\pm$ 0.01 & 0.914 $\pm$ 0.01 & 0.031 $\pm$ 0.0 & 0.026 $\pm$ 0.01 & 0.058 $\pm$ 0.02 & 0.889 $\pm$ 0.01 & 0.904 $\pm$ 0.01 \\ 
    Pretrained & Swapping & 0.027 $\pm$ 0.01 & 0.065 $\pm$ 0.02 & 0.892 $\pm$ 0.01 & 0.92 $\pm$ 0.01 & 0.01 $\pm$ 0.0 & 0.029 $\pm$ 0.01 & 0.066 $\pm$ 0.01 & 0.893 $\pm$ 0.01 & 0.921 $\pm$ 0.01 \\ \hline
    Context-debias & Default & 0.047 $\pm$ 0.02 & 0.128 $\pm$ 0.04 & 0.895 $\pm$ 0.01 & 0.931 $\pm$ 0.01 & 0.116 $\pm$ 0.01 & 0.049 $\pm$ 0.02 & 0.032 $\pm$ 0.02 & 0.868 $\pm$ 0.01 & 0.894 $\pm$ 0.01\\
    Context-debias & Scrubbing & 0.014 $\pm$ 0.01 & 0.032 $\pm$ 0.02 & 0.701 $\pm$ 0.22 & 0.874 $\pm$ 0.06 & 0.015 $\pm$ 0.01 & 0.015 $\pm$ 0.01 & 0.026 $\pm$ 0.02 & 0.698 $\pm$ 0.21 & 0.87 $\pm$ 0.06 \\
    Context-debias & Swapping  & 0.031 $\pm$ 0.01 & 0.049 $\pm$ 0.02 & 0.889 $\pm$ 0.01 & 0.915 $\pm$ 0.01 & 0.01 $\pm$ 0.0 & 0.031 $\pm$ 0.01 & 0.05 $\pm$ 0.02 & 0.888 $\pm$ 0.01 & 0.915 $\pm$ 0.01 \\ \hline
    CDA& Default & 0.055 $\pm$ 0.02 & 0.114 $\pm$ 0.03 & 0.902 $\pm$ 0.01 & 0.932 $\pm$ 0.0 & 0.115 $\pm$ 0.01 & 0.048 $\pm$ 0.01 & 0.024 $\pm$ 0.01 & 0.872 $\pm$ 0.01 & 0.899 $\pm$ 0.0 \\
    CDA & Scrubbing & 0.023 $\pm$ 0.01 & 0.053 $\pm$ 0.03 & 0.894 $\pm$ 0.01 & 0.904 $\pm$ 0.01 & 0.018 $\pm$ 0.0 & 0.022 $\pm$ 0.01 & 0.041 $\pm$ 0.02 & 0.892 $\pm$ 0.01 & 0.899 $\pm$ 0.01 \\ 
    CDA & Swapping & 0.026 $\pm$ 0.02 & 0.038 $\pm$ 0.02 & 0.891 $\pm$ 0.01 & 0.913 $\pm$ 0.01 & 0.008 $\pm$ 0.0 & 0.026 $\pm$ 0.02 & 0.036 $\pm$ 0.02 & 0.891 $\pm$ 0.01 & 0.913 $\pm$ 0.01\\ \hline
    Sent-debias & Original & 0.074 $\pm$ 0.02 & 0.108 $\pm$ 0.03 & 0.898 $\pm$ 0.01 & 0.926 $\pm$ 0.01 & 0.069 $\pm$ 0.01 & 0.041 $\pm$ 0.02 & 0.025 $\pm$ 0.01 & 0.865 $\pm$ 0.01 & 0.89 $\pm$ 0.01 \\
    Sent-debias & Scrubbing & 0.033 $\pm$ 0.01 & 0.035 $\pm$ 0.01 & 0.889 $\pm$ 0.01 & 0.909 $\pm$ 0.01 & 0.012 $\pm$ 0.0 & 0.037 $\pm$ 0.01 & 0.023 $\pm$ 0.02 & 0.885 $\pm$ 0.01 & 0.901 $\pm$ 0.01 \\
    Sent-debias & Swapping & 0.021 $\pm$ 0.02 & 0.045 $\pm$ 0.01 & 0.893 $\pm$ 0.01 & 0.916 $\pm$ 0.01 & 0.006 $\pm$ 0.0 & 0.021 $\pm$ 0.02 & 0.045 $\pm$ 0.01 & 0.893 $\pm$ 0.01 & 0.915 $\pm$ 0.01 
\\
  \end{tabular}
}
\end{table*}

\begin{table*}[tbh]
\caption{Extrinsic bias scores from Jigsaw - ALBERT-large}\label{tab:Jigsaw Albert}
\centering
\resizebox{\linewidth}{!}{%
\renewcommand{\arraystretch}{1} 
  \begin{tabular}{l|l|llll|l|llll}
    \toprule
      \hline
      \textbf{Model} & \textbf{Data} & \textbf{TPRD}  & \textbf{FPRD}& \textbf{ACC-F}&\textbf{ACC-M} & \textbf{CF} & \textbf{CF-TPRD} &
      \textbf{CF-FPRD} & \textbf{CF-ACC-F} & \textbf{CF-ACC-M} \\\hline
        \midrule
    Pretrained & Default & 0.077 $\pm$ 0.04 & 0.115 $\pm$ 0.04 & 0.894 $\pm$ 0.01 & 0.916 $\pm$ 0.01 & 0.143 $\pm$ 0.01 & 0.035 $\pm$ 0.02 & 0.039 $\pm$ 0.02 & 0.862 $\pm$ 0.01 & 0.86 $\pm$ 0.02 \\ 
    Pretrained & Scrubbing & 0.032 $\pm$ 0.02 & 0.025 $\pm$ 0.02 & 0.789 $\pm$ 0.14 & 0.819 $\pm$ 0.1 & 0.031 $\pm$ 0.01 & 0.035 $\pm$ 0.03 & 0.033 $\pm$ 0.02 & 0.783 $\pm$ 0.14 & 0.808 $\pm$ 0.1 \\ 
    Pretrained & Swapping 
& 0.037 $\pm$ 0.02 & 0.029 $\pm$ 0.01 & 0.869 $\pm$ 0.02 & 0.895 $\pm$ 0.02 & 0.012 $\pm$ 0.0 & 0.038 $\pm$ 0.02 & 0.032 $\pm$ 0.02 & 0.869 $\pm$ 0.02 & 0.895 $\pm$ 0.02 \\ \hline
    Context-debias & Default & 0.064 $\pm$ 0.02 & 0.122 $\pm$ 0.02 & 0.891 $\pm$ 0.01 & 0.92 $\pm$ 0.01 & 0.137 $\pm$ 0.01 & 0.051 $\pm$ 0.02 & 0.025 $\pm$ 0.01 & 0.857 $\pm$ 0.01 & 0.875 $\pm$ 0.01 \\
    Context-debias & Scrubbing & 0.031 $\pm$ 0.01 & 0.05 $\pm$ 0.03 & 0.885 $\pm$ 0.01 & 0.891 $\pm$ 0.01 & 0.039 $\pm$ 0.01 & 0.024 $\pm$ 0.01 & 0.041 $\pm$ 0.02 & 0.878 $\pm$ 0.01 & 0.874 $\pm$ 0.01 \\
    Context-debias & Swapping & 0.023 $\pm$ 0.01 & 0.048 $\pm$ 0.02 & 0.874 $\pm$ 0.01 & 0.904 $\pm$ 0.01 & 0.011 $\pm$ 0.0 & 0.025 $\pm$ 0.01 & 0.048 $\pm$ 0.02 & 0.875 $\pm$ 0.01 & 0.905 $\pm$ 0.01 \\ \hline
    CDA& Default & 0.078 $\pm$ 0.02 & 0.115 $\pm$ 0.03 & 0.887 $\pm$ 0.01 & 0.913 $\pm$ 0.01 & 0.139 $\pm$ 0.01 & 0.043 $\pm$ 0.02 & 0.031 $\pm$ 0.01 & 0.855 $\pm$ 0.01 & 0.865 $\pm$ 0.01 \\
    CDA & Scrubbing & 0.026 $\pm$ 0.01 & 0.052 $\pm$ 0.02 & 0.866 $\pm$ 0.02 & 0.865 $\pm$ 0.02 & 0.024 $\pm$ 0.0 & 0.021 $\pm$ 0.01 & 0.036 $\pm$ 0.02 & 0.867 $\pm$ 0.02 & 0.857 $\pm$ 0.02 \\ 
    CDA & Swapping & 0.042 $\pm$ 0.02 & 0.038 $\pm$ 0.02 & 0.866 $\pm$ 0.01 & 0.9 $\pm$ 0.01 & 0.011 $\pm$ 0.0 & 0.041 $\pm$ 0.02 & 0.037 $\pm$ 0.02 & 0.867 $\pm$ 0.01 & 0.9 $\pm$ 0.01 
\\ \hline
    Sent-debias & Original & 0.102 $\pm$ 0.05 & 0.106 $\pm$ 0.05 & 0.765 $\pm$ 0.18 & 0.87 $\pm$ 0.05 & 0.064 $\pm$ 0.03 & 0.023 $\pm$ 0.02 & 0.03 $\pm$ 0.02 & 0.736 $\pm$ 0.17 & 0.824 $\pm$ 0.03 

\\
    Sent-debias & Scrubbing & 0.011 $\pm$ 0.01 & 0.021 $\pm$ 0.02 & 0.685 $\pm$ 0.21 & 0.831 $\pm$ 0.04 & 0.009 $\pm$ 0.01 & 0.011 $\pm$ 0.01 & 0.018 $\pm$ 0.02 & 0.683 $\pm$ 0.21 & 0.825 $\pm$ 0.03 
\\
    Sent-debias & Swapping & 0.031 $\pm$ 0.02 & 0.031 $\pm$ 0.02 & 0.821 $\pm$ 0.13 & 0.88 $\pm$ 0.04 & 0.006 $\pm$ 0.0 & 0.034 $\pm$ 0.02 & 0.031 $\pm$ 0.02 & 0.818 $\pm$ 0.13 & 0.88 $\pm$ 0.04 
\\
  \end{tabular}
  }
\end{table*}

\begin{figure*}[tbh]
     \centering
               \begin{subfigure}{0.45\textwidth}
         \centering
         \includegraphics[width=\textwidth]{figures/BERT_SEAT_LPBS.png}
         \caption{BERT-large model}
         \label{fig:Bert Large intrinstic bias-2}
     \end{subfigure}
     \hfill
     \begin{subfigure}{0.45\textwidth}
         \centering
         \includegraphics[width=\textwidth]{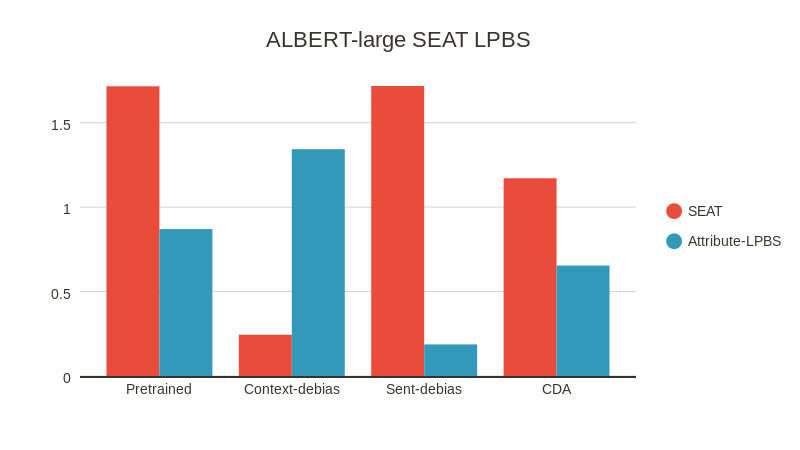}
         \caption{ALBERT-large model}
         \label{fig:Albert Large intrinsic bias}
     \end{subfigure}
\caption{Intrinsic bias scores using SEAT and LPBS. Results show inconsistencies in measuring bias between SEAT and LPBS for various mitigation strategies --- lower scores are desirable in both cases.}\label{fig:intrinsic bias}
\end{figure*}

\begin{figure*}[tbh]
     \centering
          
     \begin{subfigure}{0.45\textwidth}
         \centering
         \includegraphics[width=\textwidth]{figures/BERT_LPBS_LPBS.png}
         \label{fig:Bert LPBS Large intrinstic bias-2}
     \end{subfigure}
     \hfill
     \begin{subfigure}{0.45\textwidth}
         \centering
         \includegraphics[width=\textwidth]{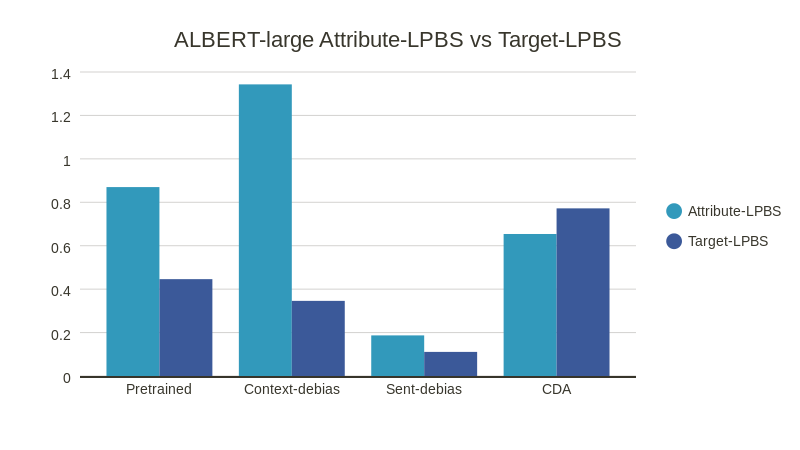}
         \label{fig:Albert LPBS Large intrinsic bias}
     \end{subfigure}
\caption{We compare LPBS in its default form (attribute-LPBS) and target-LPBS for both BERT and ALBERT. The plot shows a general positive correlation between the two metrics. In both cases, Sent-debias maintains the lowest bias score --- lower scores are desirable in both cases}\label{fig:intrinsic bias LPBS}
\end{figure*}

\begin{figure*}[tbh]
     \centering
     \begin{subfigure}{0.45\textwidth}
         \centering
         \includegraphics[width=\textwidth]{figures/bert_large_gender_detection.png}
         \label{fig:Bert Large gender detection-2}
     \end{subfigure}
     \hfill
     \begin{subfigure}{0.45\textwidth}
         \centering
         \includegraphics[width=\textwidth]{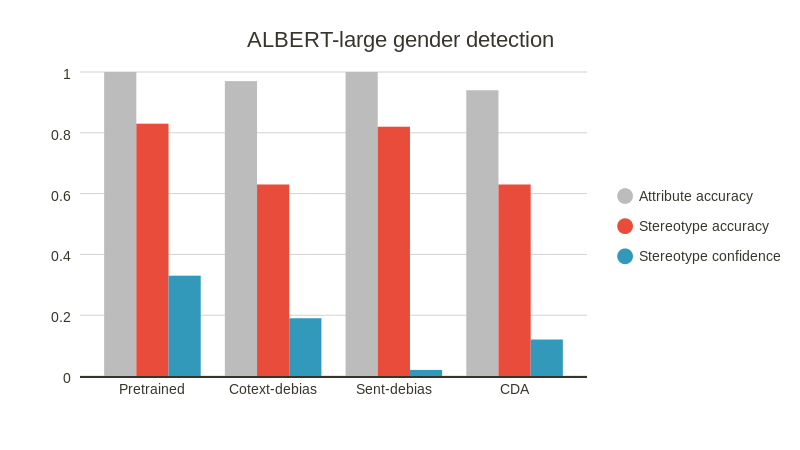}
         \label{fig:Albert Large gender detection}
     \end{subfigure} 
     \caption{Gray bars indicate the accuracy of detecting gender information in attribute terms (high scores are desirable). Red bars indicate the accuracy of detecting gender information in stereotype terms (this should ideally be 0.5 in a fair classifier, showing the inability to correctly predict the gender association of stereotype terms). Blue bars indicate the average confidence of prediction: $\frac{1}{N}\sum_i^N |bias(e_i)-0.5|$ (low scores are desirable).}
     \label{fig:Gender_detection}
\end{figure*}

\begin{figure*}
     \centering
     \begin{subfigure}[b]{0.45\textwidth}
         \centering
         \includegraphics[width=\textwidth]{figures/bias_in_bios/bias-in-bios-bert-large-TPRD.png}
         \label{fig:Bert Large Intrinsic-extrinsic TPRD}
     \end{subfigure}
     \hfill
     \begin{subfigure}[b]{0.45\textwidth}
         \centering
         \includegraphics[width=\textwidth]{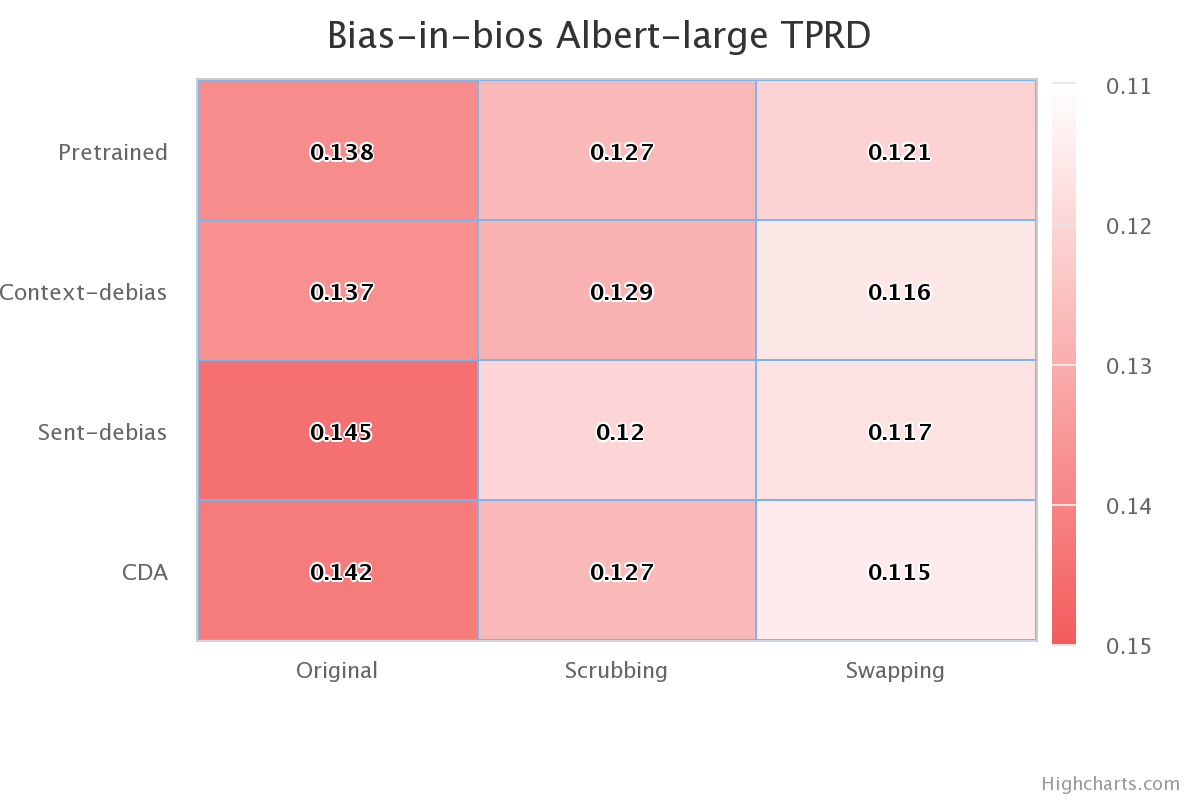}
         \label{fig:Albert Large Intrinsic-extrinsic TPRD}
     \end{subfigure}
     \begin{subfigure}[c]{0.45\textwidth}
         \centering
         \includegraphics[width=\textwidth]{figures/bias_in_bios/bias-in-bios-bert-large-CF.png}
         \label{fig:Bert Large Intrinsic-extrinsic CF}
     \end{subfigure}
     \hfill
     \begin{subfigure}[d]{0.45\textwidth}
         \centering
         \includegraphics[width=\textwidth]{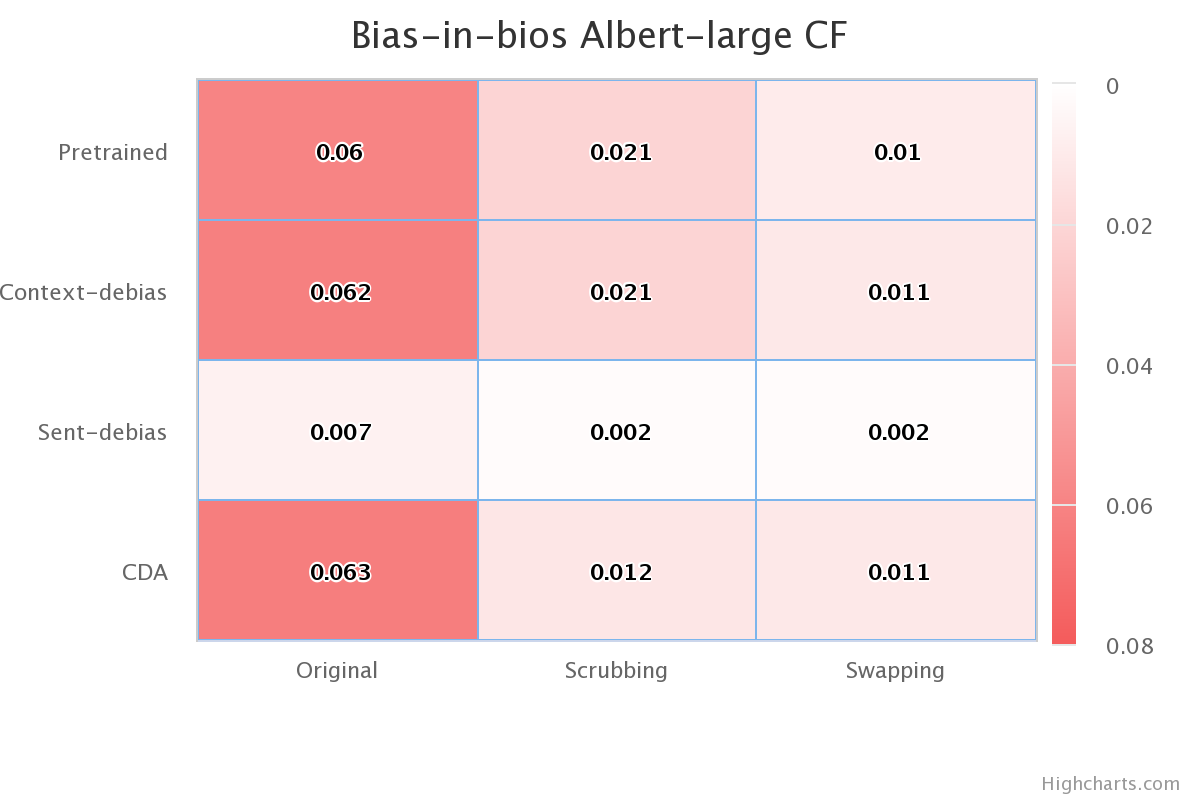}
         \label{fig:Albert Large Intrinsic-extrinsic CF}
     \end{subfigure}
     \caption{How intrinsic bias mitigation and downstream data intervention interact to influence fairness on the bias-in-bios dataset.} \label{fig:bias-in-bios Intrinsic-extrinsic bias}
\end{figure*}

\begin{figure*}
     \centering
     \begin{subfigure}[b]{0.45\textwidth}
         \centering
         \includegraphics[width=\textwidth]{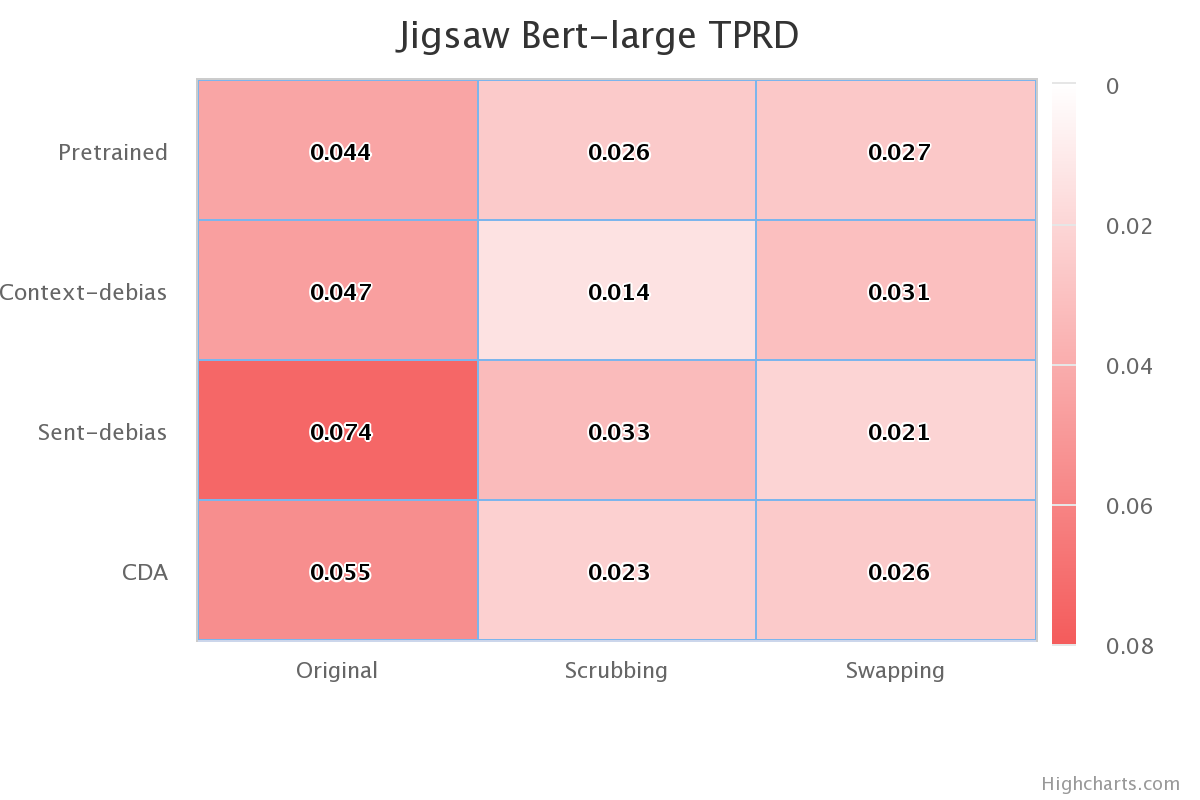}
         \label{fig:jigsaw Bert Large Intrinsic-extrinsic TPRD}
     \end{subfigure}
     \hfill
     \begin{subfigure}[b]{0.45\textwidth}
         \centering
         \includegraphics[width=\textwidth]{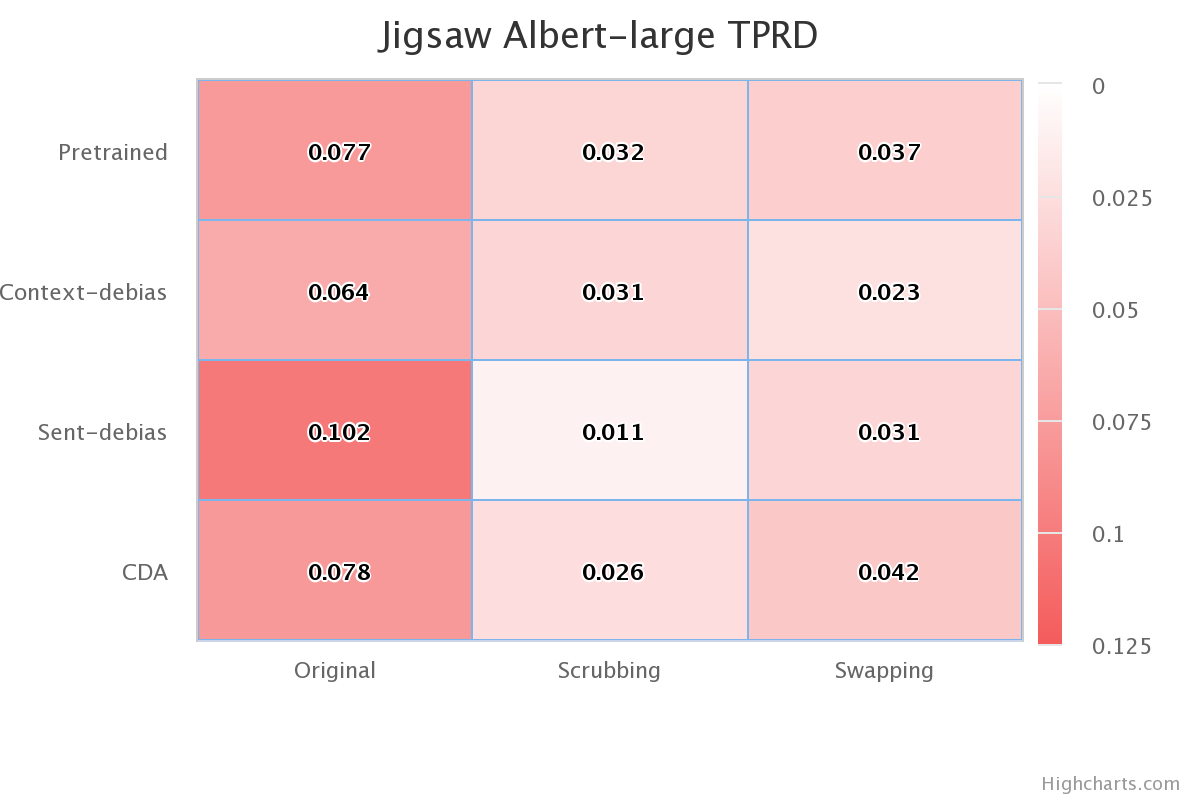}
         \label{fig:jigsaw Albert Large Intrinsic-extrinsic TPRD}
     \end{subfigure}
     \begin{subfigure}[c]{0.45\textwidth}
         \centering
         \includegraphics[width=\textwidth]{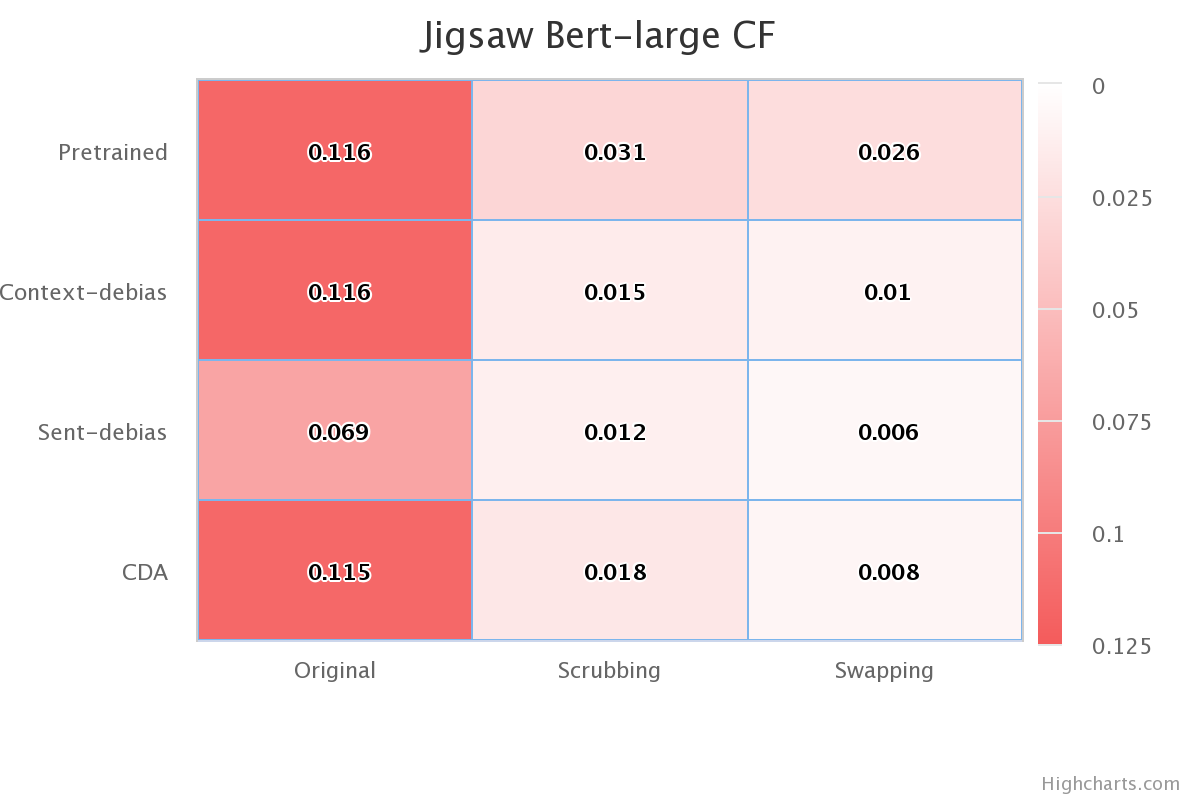}
         \label{fig:jigsaw Bert Large Intrinsic-extrinsic CF}
     \end{subfigure}
     \hfill
     \begin{subfigure}[d]{0.45\textwidth}
         \centering
         \includegraphics[width=\textwidth]{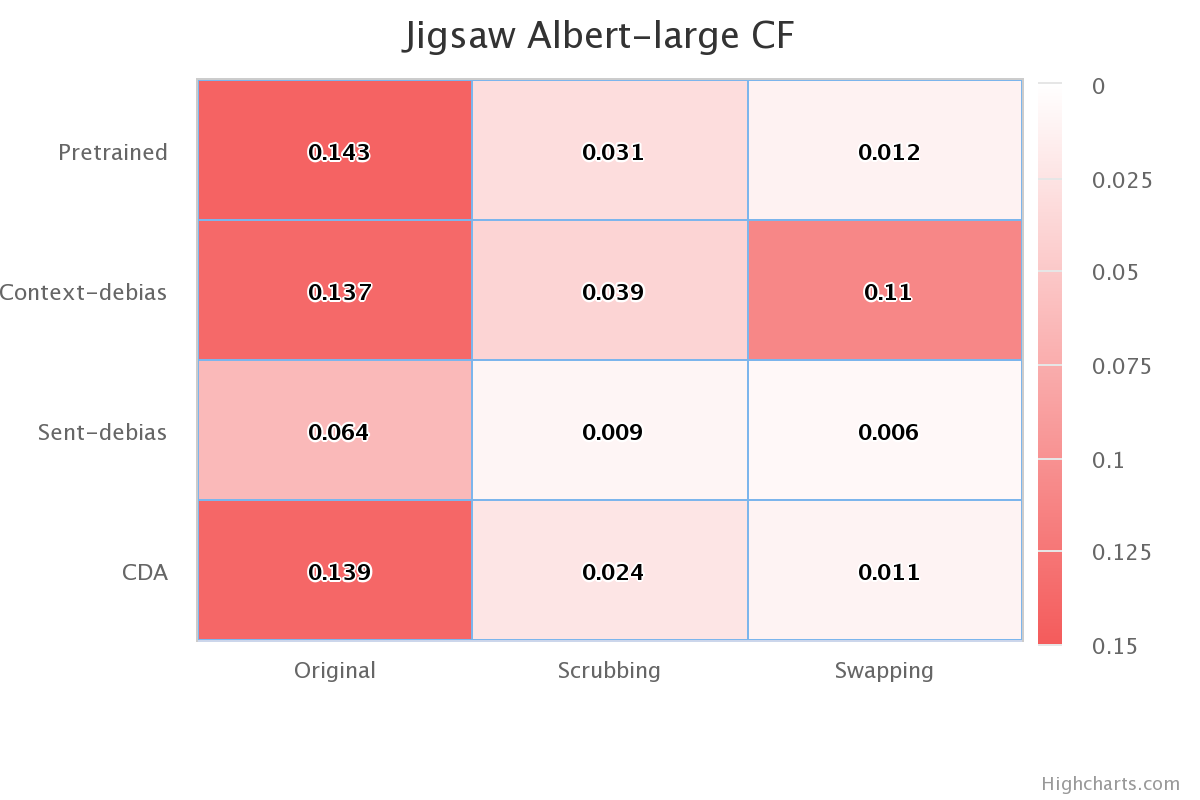}
         \label{fig:jigsaw Albert Large Intrinsic-extrinsic CF}
     \end{subfigure}
     \caption{How intrinsic mitigation and downstream data intervention interact to influence counterfactual fairness on the jigsaw data. }\label{fig:jigsaw extrinsic-Intrinsic bias}
\end{figure*}

\begin{figure*}[tbh]
     \centering
     \begin{subfigure}[b]{0.48\textwidth}
         \centering
         \includegraphics[width=\textwidth]{figures/tpr-v-cf-bert-large.png}
         \label{fig:TPR-CF-Bert-Large-2}
     \end{subfigure}
     \hfill
     \begin{subfigure}[b]{0.48\textwidth}
         \centering
         \includegraphics[width=\textwidth]{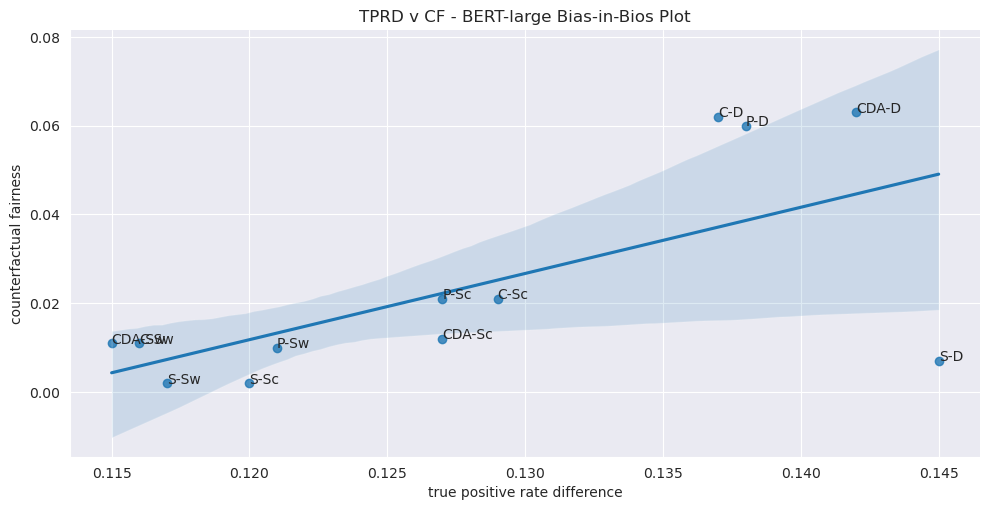}
         \label{fig:TPR-CF-Albert-Large}
     \end{subfigure}
     \caption{Relationship between TPRD and CF based on results from the Bias-in-Bios dataset.
}
    \label{fig:TPRD-CF}
\end{figure*}

\end{document}